\documentclass[journal]{IEEEtran}      

\IEEEoverridecommandlockouts

\usepackage{graphicx} 
\usepackage{tensor}
\usepackage[font=footnotesize]{caption}
\usepackage{cite}
\usepackage[font=footnotesize]{subcaption}
\usepackage{booktabs}
\usepackage{times} 
\usepackage{bm}
\usepackage{amsmath,amssymb,mathrsfs,mathdesign,upgreek,dsfont,gensymb}
\usepackage[ruled,vlined]{algorithm2e}
\usepackage{verbatim}
\usepackage{diagbox}
\usepackage[hidelinks]{hyperref}
\hypersetup{
   linkcolor=blue,
   breaklinks=true,  
   colorlinks=true, 
   citecolor=blue,
   urlcolor=blue
}

\usepackage{cleveref}
\usepackage{soul}
\usepackage[multiple]{footmisc}
\usepackage{multirow}
\usepackage{threeparttable}
\usepackage{enumitem}
\usepackage{color}
\usepackage{xspace}

\graphicspath{{./figures/}}

\DeclareCaptionSubType*[Alph]{table}
\DeclareCaptionLabelFormat{mystyle}{Table~\bothIfFirst{#1}{ ̃}#2}

\def\code#1{\texttt{#1}}
\DeclareMathOperator*{\argmin}{arg\!\min}

\DeclareMathOperator*{\Log}{log}
\newcommand*\diff{\mathop{}\!\mathrm{d}}

\newcommand \facto{FACTO\xspace}
\newcommand \figref{Figure~\ref}
\newcommand \tabref{Table~\ref}
\newcommand \secref{Section~\ref}
\newcommand \algref{Algorithm~\ref}

\title{
\facto: Function-space Adaptive Constrained Trajectory Optimization for Robotic Manipulators 
}

\author{
Yichang Feng$^{1}$, Xiao Liang$^{2,*}$, and Minghui Zheng$^{1,*}$
\thanks{$^{1}$Yichang Feng and Minghui Zheng are with the J. Mike Walker '66 Department of Mechanical Engineering, Texas A\&M University, College Station, TX 77843, USA (\tt\footnotesize e-mail: yichangfeng@tamu.edu; mhzheng@tamu.edu).}
\thanks{$^{2}$Xiao Liang is with the Zachry Department of Civil and Environmental
Engineering, Texas A\&M University, College Station, TX 77843 USA (\tt\footnotesize e-mail:
xliang@tamu.edu).}
\thanks{$^*$ Corresponding Authors.}
\thanks{The source code for this work will be made available in a public repository upon acceptance of the manuscript.}
}

\begin{document}

\maketitle
\thispagestyle{empty}
\pagestyle{empty}

\begin{abstract}

This paper introduces \textbf{Function-space Adaptive Constrained Trajectory Optimization (\facto)}, a new trajectory optimization algorithm for both single- and multi-arm manipulators. Trajectory representations are parameterized as linear combinations of orthogonal basis functions, and optimization is performed directly in the coefficient space. The constrained problem formulation consists of both an objective functional and a finite-dimensional objective defined over truncated coefficients. To address nonlinearity, \facto\ uses a Gauss-Newton approximation with exponential moving averaging, yielding a smoothed quadratic subproblem. Trajectory-wide constraints are addressed using coefficient-space mappings, and an adaptive constrained update using the Levenberg-Marquardt algorithm is performed in the null space of active constraints.
Comparisons with optimization-based planners (CHOMP, TrajOpt, GPMP2) and sampling-based planners (RRT-Connect, RRT*, PRM) show the improved solution quality and feasibility, especially in constrained single- and multi-arm scenarios. The experimental evaluation of \facto on Franka robots verifies the feasibility of deployment. The supplemental video is available via this \href{https://zh.engr.tamu.edu/wp-content/uploads/sites/310/2026/02/FACTO_SupplementalVideo_compr.mp4}{link}.

\end{abstract}

\section{Introduction} 

Planning collision-free and dynamically feasible trajectories that satisfy specific task constraints with \emph{time-continuous} feasibility remains difficult in cluttered workspaces. In many manipulation tasks, such as pick-and-place, door-handle grasping, or multi-arm cooperation, only a few localized adjustments to the motion trajectory are needed to satisfy task constraints and avoid obstacles. Most existing approaches discretize the continuous-time trajectory into a series of discrete waypoints and usually focus on finding a feasible solution in the discrete trajectory space. To guarantee time-continuous feasibility, some optimization-based motion planners \cite{zucker_chomp_2013, schulman_motion_2014, mukadam_continuous-time_2018} often increase the number of discrete waypoints, and some sampling-based motion planners \cite{kavraki_analysis_1998, kuffner_rrt-connect_2000} often increase the sampling density in the configuration space.

Waypoint-based optimization methods, such as CHOMP \cite{zucker_chomp_2013},  TrajOpt \cite{schulman_motion_2014}, and GPMP2 \cite{mukadam_continuous-time_2018}, represent the trajectory by a series of support states. Although waypoint sparsity can reduce computational cost by reducing the number of dimensions, it compromises time-continuous safety because the feasibility of the transition between adjacent waypoints is not certified. To address this problem, some propose the up-sampling or convex-hull sweep methods for intermediate-point interpretation, but these methods always require additional collision checks and Jacobian calculations. Others find smooth trajectories in early iterations with a good initial guess. Moreover, the trajectory-wide equality constraints from the task requirements scale with the number of support states, thereby increasing the linearization and checking costs when continuous-time guarantees are required.

Sampling-based planners, like PRM \cite{kavraki_analysis_1998}, RRT \cite{lavalle_rapidly-exploring_1998}, RRT-connect \cite{kuffner_rrt-connect_2000}, and their asymptotically-optimal variants \cite{karaman_sampling-based_2011, otte_rrtx_2016}, are probabilistically complete and adequate at exploring feasible configuration-space regions. However, they and their informed or heuristic variants \cite{rajendran_context-dependent_2019, gammell_batch_2020} cannot meet the task requirements. In this way, the implicit manifold configuration space (IMACS \cite{kingston_exploring_2019}), an unifying framework for constrained sampling, decouples the planner algorithms from the constraint-adherence methods \cite{jaillet_path_2013, kim_tangent_2016}, and shows that guarantees are preserved. However, a large amount of computation is required for the addition-constraints check and the subsequent trajectory-optimization stage. 

\begin{figure}
    \centering
    \includegraphics[width=.95\linewidth]{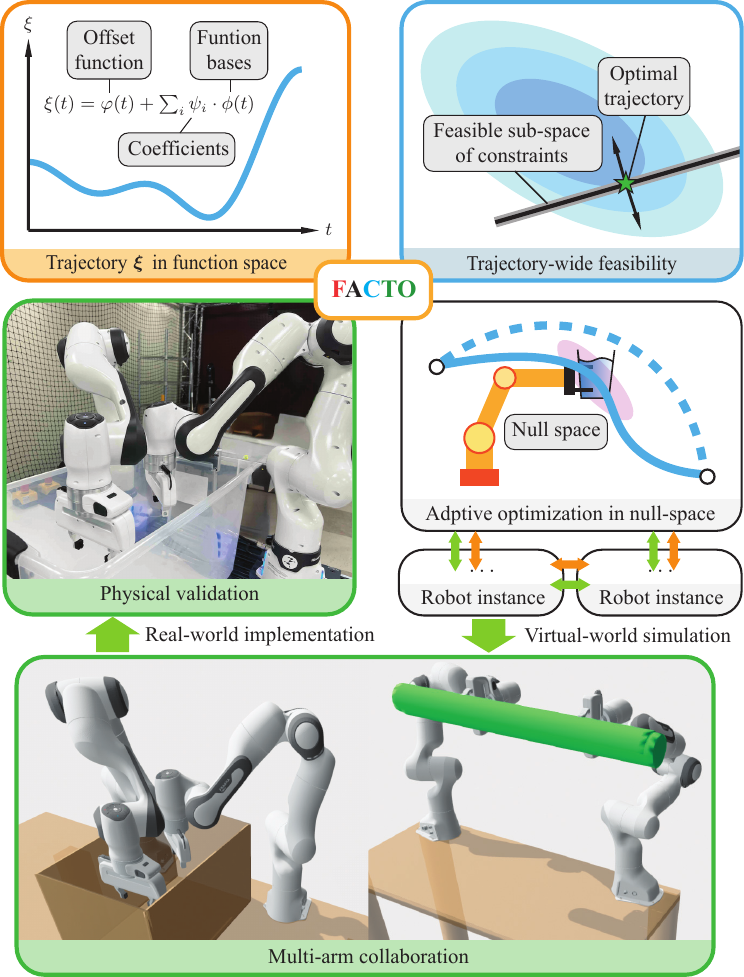}
    \caption{
    Overview of the \facto framework. 
    (Top left) Joint trajectories are represented in a coefficient space using an offset function and a set of orthogonal basis functions. 
    (Top right) Equality and inequality constraints define a feasible subspace in coefficient space, within which the update step is selected. 
    (Bottom right) Active constraints are enforced through null-space projection, enabling adaptive optimization while preserving feasibility. 
    (Bottom left) In multi-arm settings, robot instances exchange only the information required for self-collision checking and coordination/collaboration tasks. 
    Overall, these components form a complete \facto pipeline for generating continuous-time, constraint-satisfying trajectories in both single- and multi-robot systems.
    }
    \label{fig:facto}
\end{figure}

We propose \facto, Function-space Adaptive Constrained Trajectory Optimization. 
We formulate trajectory optimization directly in a \emph{function space} and parameterize a time-continuous trajectory in a coefficient space, shown in \figref{fig:facto}. It means that each joint trajectory is expressed as a linear combination of \emph{orthogonal basis functions} rather than serial waypoints in an execution time period $[0,T]$. The optimization variables are instead the corresponding coefficients of the basis functions. 
Since most local-minima cases exhibit a sudden change in the sub-trajectory, introducing high-frequency noise, the truncated series of orthogonal basis functions can use a low bandwidth to ensure time-continuous safety. Since the number of linearly independent constraints is smaller than the number of coefficients to ensure the solution's existence, the number of trajectory-wide constraint checks is also significantly reduced. 

Equality constraints, like boundary conditions and task frames, are enforced \emph{analytically} by projecting search directions onto the \emph{null space} of the active equality. It ensures feasibility by construction and shrinks the search space. Trajectory-wide inequality constraints, such as joint limits, are handled using a compact \emph{active set}, since the coefficients are global; updating a few active constraints refines the entire trajectory. 
Considering discontinuity and nonlinearity of the objective function, caused by the sampled collision check, the quadratic programming problem is constructed by the moving average of the gradient and curvature information. The overall optimization employs the Gauss–Newton method with adaptive trust regions to ensure a more robust and efficient convergence. 

The challenges of multi-robot trajectory optimization stem from the higher degrees of freedom, self-collision checking, and the construction of constraints for collaborative tasks. \facto fully decouples the multi-robot system into individual robots and optimizes their trajectories independently. The cross-check matrix is adopted and simplified to ensure self-collision-free. The collaborative tasks are decentralized, and virtual task constraints are allocated to each robot in the fully decoupled multi-robot system. 

\paragraph*{Our contributions can be summarized as follows}
\begin{itemize}[itemindent=2\parindent, leftmargin=0pt] 
  \item \textbf{Optimization in a truncated function space.}
  We formulate collision- and task-aware trajectory optimization directly over truncated coefficients of orthogonal basis functions, yielding low-dimensional updates with trajectory-wide effect and smooth, low-bandwidth solutions.

  \item \textbf{Equality feasibility.}
  We enforce the boundary and linearized task constraints via null-space projection, yielding a reduced-space SQP step that maintains equality feasibility throughout the iterations.

  \item \textbf{Sparse trajectory-wide limit enforcement.} 
  We maintain a small and active set of inequality constraints and enforce joint limits via coefficient-space bounds evaluated at a few critical time nodes, thereby avoiding dense sampling. Considering the feasibility of the velocity and torque, we scale the execution time of each optimally parameterized trajectory. 

  \item \textbf{Robust SQP updates.}
  We introduce an EMA-regularized Gauss-Newton curvature and an adaptive trust-region scheme that stabilizes convergence for highly nonconvex and constrained optimization problems.

  \item \textbf{Scalable multi-robot optimization.}
  We exploit the block structure of multi-robot objectives and constraints to decouple the multi-robot system into per-robot subproblems, while handling collaborative task constraints through a decentralized coordination scheme.
\end{itemize}

In summary, \facto combines the efficiency of trajectory parameterization in the coefficient space with a supportive framework for constraint satisfaction and optimization over the entire trajectory. This is achieved by overcoming the key challenges of waypoint-based methods, such as CHOMP, GPMP2, and TrajOpt, as well as sampling-based motion planners, such as RRT-Connect, IMACS, RRT*, and PRM. Our benchmarking experiment, with 800 unconstrained and 200 constrained tasks on a single-arm panda robot, and 40 unconstrained and 40 constrained tasks on a dual-arm panda robot, demonstrates the outstanding efficiency and effectiveness of \facto. Additionally, our physical experiment validates \facto's applicability across various tasks, including single-arm manipulation and dual-arm coordination/collaboration.

\section{Related Works}

The previous work is reviewed in this section using three key topics: \emph{trajectory representation}, \emph{continuous-time (trajectory-wide) constraint enforcement}, and \emph{optimizer globalization}. We start with waypoint models and their limitations in time-continuous settings. Then we cover function-space parameterizations that motivate \facto. Finally, we review sampling-based and learning-augmented planners. 

\subsection{Waypoint Parameterizations and Continuous-Time Gap}

Most optimization-based motion planners numerically optimize the discrete \emph{waypoints}. CHOMP \cite{zucker_chomp_2013} uses the functional gradient for motion updates. Meanwhile, TrajOpt \cite{schulman_motion_2014} convexifies the objective and adjusts trust regions for fast convergence. These methods work well on moderate problems but inherit a structural limitation: sparse waypoints break \emph{time-continuous} guarantees because local collision-avoidance updates omit information between waypoints. To address these issues, GPMP/GPMP2 \cite{mukadam_continuous-time_2018}
uses dense up-sampling, increasing collision checks and Jacobian evaluations, and TrajOpt uses convex-hull sweeps, assuming sufficient smoothness but sensitive to initial guess. Though GPMP-GRAPH \cite{huang_motion_2017}, a batch optimizer, and GOMP \cite{ichnowski_deep_2020}, integrating deep learning, extend the scope, the discretization burden remains. 
The time-continuous issue is also faced by the SQP- and MPC-style formulations for avoidance and coordination \cite{woolfrey_predictive_2021, mao_real-time_2019, agravante_human-humanoid_2019, mo_kind_2020, kabir_generation_2021} when dealing with trajectory-wide inequalities. 

Recent work has pushed waypoint pipelines toward better practice. Warm-start and reuse across related problems can improve the solution feasibility with high speed \cite{lembono_memory_2020}, and multimodal optimization seeks diverse solutions in the presence of nonconvexity \cite{osa_multimodal_2020}. Hybrid schemes also interleave sampling with local optimization to quickly repair or refine candidates in clutter \cite{kuntz_fast_2020}. 
These works improve the reliability and efficiency, though the continuous-time guarantee still requires dense checking or conservative bounds. 

However, these waypoint models still make time-continuous safety and constraint handling \emph{scale with sampling density}, meaning that improving one may worsen runtime. Sometimes, their local effects on the entire trajectory also lead to planning failure, no matter how many waypoints are deployed. 
These motivate trajectory parameterization in a space where \emph{smoothness} and \emph{continuous-time reasoning} are intrinsic.

\subsection{Function-Space Parameterizations}

An alternative is to optimize over a function space defined by trajectory parameterizations, such as Gaussian processes or basis functions. Such parameterizations reduce the number of optimization variables and propagate local changes globally. Within the GP family \cite{mukadam_continuous-time_2018, huang_motion_2017}, recent efforts tend to increase modeling and inference flexibility beyond classical GPMP2 by variational unifications \cite{cosier_unifying_2024}, differentiable formulations \cite{bhardwaj_differentiable_2020}, or constrained inference procedures \cite{li_constrained_2025}. In parallel, a Fourier-based parameterization is proposed to produce low-bandwidth motions \cite{feng_optimal_2023}, offering a practical way to achieve sparse sampling without losing expressiveness.

\facto follows this line to propose a generic framework with two elements: (i) an \emph{orthogonal} basis, such as trigonometric, Chebyshev/Legendre, compact splines/wavelets, that yields closed-form derivatives and well-conditioned updates; and (ii) \emph{coefficient-space} enforcement of trajectory-wide constraints via convex envelopes, removing dense inter-sample checks.  

\subsection{Mechanisms for Trajectory-Wide Safety}

Prior works address the between-waypoints gap by (i) up-sampling, proposed by GPMP2 \cite{dong_motion_2016} to insert dense intermediate states, or by (ii) convex-hull sweeps, proposed by TrajOpt \cite{schulman_motion_2014}, which requires high smoothness and good initializations. 

Some recent works improve the efficiency of safety verification without compromising the guarantees. For example, the learned proxy collision checking aims to replace some geometric queries with fast predictors \cite{das_learning-based_2020}, while safe planning under uncertainty considers constraints imposed by partial observability and risk \cite{hibbard_safely_2023}. 

Our function-space method instead expresses \emph{continuous-time} envelopes in coefficient space. Joint limits then become a few convex hulls despite the sampling density, and boundary constraints are enforced via null-space projection. 
This shift underlies \facto's ability to certify continuous-time properties without exploding the number of constraints.

\subsection{Sampling-Based Planning and Manifold Variants}

Sampling-based planners build trees or roadmaps to explore feasible regions: RRT, RRT*, RRT-Connect, LQR-RRT*, RRTX, NoD-RRT, and BIT* improve quality, bidirectionality, control-awareness, and replanning \cite{lavalle_rapidly-exploring_1998, karaman_sampling-based_2011, kuffner_rrt-connect_2000, perez_lqr-rrt_2012,otte_rrtx_2016, li_neural_2018,gammell_batch_2020}, with memory and efficiency refinements (RRT-FN \cite{adiyatov_rapidly-exploring_2013}, CODE3 \cite{rajendran_context-dependent_2019}). PRM variants target dynamics, multi-arm, or workspace-guided sampling \cite{kavraki_randomized_1994, kavraki_probabilistic_1996, hofmann_reactive_2015, dai_fast-reactive_2021, logothetis_motion_2019, luna_scalable_2020}. Recent methods, such as tensorized global planning \cite{le_global_2025}, continue to expand the algorithmic design space.
Atlas, tangent-bundle, and IMACS methods \cite{jaillet_path_2013, kim_tangent_2016, kingston_exploring_2019} use local projections to remain on constraint manifolds. Modern comparative surveys further clarify the empirical tradeoffs among these families across dimensions, constraints, and environmental structure \cite{orthey_sampling-based_2024}. 

Though these approaches are probabilistically complete, graph or tree construction entails a large amount of collision checks and yields non-smooth paths. And optimization is still required to satisfy continuous-time constraints, which \facto's function-space optimizer directly targets. In practice, sampling and optimization are increasingly composed in layered stacks. \cite{stouraitis_online_2020} proposes hybrid bilevel formulations for collaborative manipulation. \cite{koptev_reactive_2024} introduces reactive motion-generation modules that blend optimization with sampling.

\subsection{Learning-Augmented Planning}

Learning augments both optimization- and sampling-based planners through warm starts, cost shaping, collision predictors, and motion primitives. Such priors can accelerate convergence or bias exploration, but continuous-time constraints are not guaranteed. Recent methods include learning to accelerate grasp-optimized planning \cite{ichnowski_deep_2020} and proxy collision detection \cite{das_learning-based_2020}. In broader planning stacks, task-and-motion integration frameworks combine symbolic search with blackbox sampling components \cite{garrett_pddlstream_2020}. For manipulation settings related to our tasks, dual-arm coordination and twisting-like behaviors have also been studied in contemporary systems \cite{lin_twisting_2024, ren_enabling_2024}. 

\facto, as a numerical trajectory optimizer, proposes a framework for continuous-time guarantees provided by analytic equalities and coefficient-space envelopes. It does not integrate any learning-based technique, which makes it much more reliable, efficient, and easier to implement.

\subsection{Solver Backbones and Globalization}

Deterministic solvers span Newton/quasi-Newton (LM, BFGS \cite{nocedal_numerical_2006, levenberg_method_1944, sugihara_solvability-unconcerned_2011, dong_motion_2016, badreddine_sequential_2014, curtis_bfgs-sqp_2017}) and trust-region methods \cite{conn_trust-region_2000, schulman_trust_2015}, with operator-splitting QP backends such as OSQP \cite{stellato_osqp_2020} and ADMM \cite{boyd_distributed_2011}. Stochastic optimizers (SGD, Adam, RMSProp, AdaGrad \cite{bottou_-line_1999,kingma_adam_2014, geoffrey_neural_2012, duchi_adaptive_2011}) and trajectory perturbation schemes (STOMP \cite{kalakrishnan_stomp_2011} and variants \cite{petrovic_stochastic_2019, osa_multimodal_2020, hibbard_safely_2023, feng_incrementally_2022}) reduce sensitivity to local minima at the expense of many rollouts. 

\facto employs a Gauss–Newton backbone with adaptive trust regions and curvature smoothing, using the \emph{low-dimensional and orthogonal} coefficientization to maintain compact and stable convergence. Recent whole-body and mobile-manipulation planners further highlight the importance of solver globalization and stable real-time updates \cite{wu_real-time_2024}.

\subsection*{Summary} 
Waypoint discretization and sampling link continuous-time guarantees with the sampling density and collision-check volume. Function-space parameterizations of \facto address these concerns by (i) optimizing a small set of smooth basis coefficients, (ii) enforcing equalities analytically via null-space projections, and (iii) imposing trajectory-wide inequalities as compact coefficient-space constraints.

\section{\facto: Overview}
\facto basically treats the constrained trajectory optimization problem as a sequential quadratic programming problem over the coefficients in the null space of the constraints, which parameterize a trajectory using serial basis functions.  Therefore, \secref{sec:traj_rep_function_space} first explains how we parameterize the trajectory in a finite coefficient space. Then, Sections \ref{sec:prob_formu}-\ref{sec:constraint} detail how the objective function and its constraints are constructed in the coefficient space. Finally, \secref{sec:facto} shows how \facto performs the null-space trajectory optimization using a convexified objective and linearized constraints, including how we scale the minimal execution time.

\section{\facto: Trajectory Parameterization}
\label{sec:traj_rep_function_space}

\subsection{Orthogonal Basis for Trajectory Parameterization}
\label{sec:orthobase_traj}
Most motion planners encode a trajectory as \emph{discrete waypoints} from the start $\bm\theta_0$ to the goal $\bm\theta_g$,
\begin{equation} \label{eq:dt-traj}
    \bm{\xi} \;=\; [\bm \theta_{0}^{\top}, \bm \theta_{1}^{\top}, \dots, \bm \theta_{T}^{\top}]^{\top},
\end{equation}
but sparse supports can miss between-sample collisions and induce discontinuities in functional gradients \cite{zucker_chomp_2013}. Densification \cite{mukadam_continuous-time_2018} improves coverage at the expense of numerous additional collision checks and Jacobian evaluations.  

To maintain sparsity while ensuring time-continuity, the trajectory parameterized by a linear combination of time-continuous basis functions offers a promising approach. In this formula, trajectory optimization is achieved by modifying the coefficients of their corresponding basis functions rather than the discrete waypoints. It means that the behaviors of collision avoidance and the task execution can have a trajectory-wide rather than waypoint-wise effect. In this way, we formulate trajectory optimization as the problem of finding optimal coefficients that parameterize a function subspace via serial truncated basis functions.  Each joint trajectory is expressed in an \emph{orthogonal} basis over the time interval $[0,T]$. Let $\{\phi_n(t)\}_{n=0}^{N}$ be orthogonal basis functions. 
By stacking $\bm\phi(t) : = [\phi_0(t),\dots,\phi_N(t)]^\top\!\in\!\mathbb{R}^{N+1}$ and collecting coefficients in $\bm \psi \in \mathbb{R}^{N+1}$, the single-DoF trajectory becomes
\begin{equation} \label{eq:basis0}
    \xi(t) \;=\; \bm\phi(t)^\top \bm\psi + \varphi(t), 
\end{equation}
where a low-order \emph{boundary lift} $\varphi(t)$ is adopted for the non-periodic motion to enforce $[\xi(0), \dot{\xi}(0)]= [\theta_0, \dot{\theta}_0]$ and $[\xi(T), \dot{\xi}(T)]= [\theta_g, \dot{\theta}_g]$. It decouples endpoint feasibility from the coefficients while preserving orthogonality. 

Considering an $M$-DoF manipulator, this paper uses a vectorized form $\bm \psi=\mathrm{vec}([\bm \psi_1^\top,\dots,\bm \psi_M^\top]^\top) \!\in\!\mathbb{R}^{M(N+1)}$, where each element stacks the trajectory coefficients of the $m$-th joint:  $\bm\psi_m = [\psi_{m,0},\dots, \psi_{m,n},\dots,\psi_{m,N}] \in \mathbb{R}^{N+1}$. In correspondence, the high-dimensional orthogonal basis function can be formulated as
\begin{equation}
    \bm\Phi(t)\ =\ \mathbf I_{M \times M} \otimes \bm\phi(t)^\top \in \mathbb{R}^{M\times M(N+1)},
\end{equation}
such that the $M$-DoF trajectory and its velocity and acceleration can be expressed as
\begin{gather}
    \bm\xi(t)= \bm{\varphi}(t) + \bm\Phi(t)\bm \psi, \label{eq:traj} \\
    \dot{\bm\xi}(t) = \dot{\bm{\varphi}}(t) + \dot{\bm\Phi}(t)\bm \psi, \label{eq:traj_vel}\\
    \ddot{\bm\xi}(t) = \ddot{\bm{\varphi}}(t) + \ddot{\bm\Phi}(t)\bm \psi. \label{eq:traj_acc}
\end{gather}
It indicates that velocity or acceleration penalties and dynamics constraints couple \emph{linearly} to $\bm \psi$, enabling efficient Gauss–Newton steps and analytic null-space projections for equality feasibility, as emphasized in the introduction.

\subsection{Function-Space Optimization with Orthogonal Basis}
\label{sec:coef_space_generic}
Let $\bm\Theta \subset \mathbb{R}^{M}$ be a field of joint configuration and  $\mathcal{T} \subseteq [0, T]$ be any subset of the time interval. 
We define a trajectory/function set $\bm \Xi \ := \{ \bm\xi:  \mathcal{T} \rightarrow \bm\Theta \}$. Since $\langle \cdot, \cdot \rangle_{w}$ with a positive function $w(t)$ is equivalent to the usual norm in $L^2$ space, we can say that $(\bm\Xi, +, \langle \cdot, \cdot \rangle_{w})$ is a Hilbert space. 

This study assumes the trajectories of each joint are linearly independent ${\bm \xi}  = \bigoplus_{m = 1}^{M} {\xi}_{m}$, meaning that the trajectory optimality of each joint can be analyzed separately. Now, suppose there exists a candidate optimal trajectory ${\xi}_m^\ast(t)$. Due to the orthogonality, the optimum can be yielded onto the span of $\{\phi_n\}_{n=0}^N$: 
\begin{equation}
     ({\xi}_m^\ast - \varphi_m)(t) \;\in\; \mathrm{span}\{\phi_n(t)\}_{n=0}^{N}, \quad 
     {\bm \xi}^\ast  = \bm\varphi + \bigoplus_{m = 1}^{M} {\xi}_{m}^\ast. 
\end{equation}
And the coefficient, a.k.a. the coordinates of the $n$-th basis function for the $m$-th joint, can be calculated by
\begin{equation} \label{eq:traj_to_coeff}
    \begin{aligned}
        \psi_{m,n}^\ast 
        &\;=\; \frac{\langle {\xi}_m^\ast - \varphi_m,\phi_n\rangle_{w}}{\|\phi_n\|_{w}} \\
        &\;=\; \frac{1}{\|\phi_n\|_{w}}\!\int_{0}^{T}\! w\,({\xi}_m^\ast - \varphi_m) \,\phi_n(t)\,\mathrm dt.
    \end{aligned}
\end{equation}
Then, the $m$-th joint optimum can be approximately expressed by the linear combination of $\{\phi_n\}_{n=0}^N$, and the equation is set up when the number of bases goes to infinity:
\begin{equation} \label{eq:coeff_to_traj}
    {\xi}_m^\ast(t) = \varphi_m(t) + \lim_{N \rightarrow \infty} \sum_{n = 0}^{N} \psi_{m, n}^\ast \, \phi_n(t). 
\end{equation}

\section{\facto: Objective Formulation}
\label{sec:prob_formu}

The above section gives the intuitions behind the trajectory parameterization and optimization in the coefficient space. \eqref{eq:traj_to_coeff} and \eqref{eq:coeff_to_traj} show how the optimal trajectory $\hat{\bm\xi}$ is expressed in a function subspace spanned by the basis functions, parameterized by a coefficient vector $\bm\psi \in \mathbb{R}^{M(N+1)}$. From this, the decision variables this study aims to optimize shift from the trajectory $\bm\xi$ to the \emph{coefficients} $\bm \psi$. Then, the constrained objective functional becomes
\begin{equation} \label{eq:obj0}
    \begin{aligned}
        \min_{\bm \psi}\quad &\mathcal{F}(\bm \psi)
        \;=\; \int_{0}^{T} f\!\big(\bm \psi, \mathbf{\Phi}(t)\big)\ \mathrm dt \\
        \text{s.t.}\quad &
        \bm{\mathcal{H}}_\text{eq}(\bm \psi)=\bm 0,\qquad
        \bm{\mathcal{H}}_\text{ieq}(\bm \psi)\le \bm 0,
    \end{aligned}
\end{equation}
where $f[\cdot]$ is nonnegative and $\mathcal{F}$, designed for the trajectory smoothness and safety, is continuously differentiable,  $\bm{\mathcal{H}}_\text{eq}(\bm \psi)$ and $\bm{\mathcal{H}}_\text{ieq}(\bm \psi)$ denote the equality and inequality constraints, respectively, under the consideration of boundary conditions, task requirements, joint limits, etc. 

As explained in \secref{sec:orthobase_traj}, one main advantage of adjusting the coefficients $\bm\psi$ rather than the trajectory $\bm\xi$ itself is the trajectory-wide effect of local collision avoidance, especially when the coefficients are truncated. That is because the truncated coefficients, in correspondence to finite bases $\{\bm\phi_n(t)\}_{n = 1}^{N}$, have a low-band impact when the trajectory is treated as a generalized multi-dimensional single. The low-band method can filter out high-frequency noise, leading to local minima where time-continuous safety is not ensured.  In this way, the sub-space consisting of the truncated bases $\{\bm\phi_n(t)\}_{n = 1}^{N}$ is no longer complete. Therefore, the objective functional of \eqref{eq:obj0} becomes an objective function with a finite-dimensional variable $\bm\Psi$, and the following parts will specify how the objective function, working objective, and its convexified version, as well as constraints, are constructed. 

In this paper, \textit{trajectory-wide} refers to the continuous-time parameterization: trajectories are represented as smooth functions over $[0,T]$, and any coefficient update affects the entire time interval. In practice, joint limits, obstacle avoidance, and task constraints are enforced via collocation at a small set of critical time nodes rather than through dense sampling.

\subsection{Objective Functional}\label{sec:objective}
This study adopts the objective formula of \cite{zucker_chomp_2013, schulman_motion_2014, mukadam_continuous-time_2018} to balance between the smoothness and obstacle avoidance:
\begin{equation}
    \mathcal{F}(\bm \psi) 
    = \varrho\,\mathcal{F}_{\mathrm{smooth}}(\bm \psi)\;+\;\mathcal{F}_{\mathrm{obs}}(\bm \psi) 
    = \int_{t=0}^{T} f(\bm\psi, \bm\Phi) \diff t. 
\end{equation}

Unlike \cite{zucker_chomp_2013, mukadam_continuous-time_2018},  which utilizes the Euler-Lagrangian equation to derive the objective functional gradient:
\begin{equation} \label{eq:functinal_grad}
    \bar{\nabla}_{\bm \xi}\mathcal{F} = \frac{\partial f}{\partial \bm\xi} - \frac{\diff}{\diff t}\frac{\partial f}{\partial \dot{\bm\xi}}, 
\end{equation}
this study derives the objective function gradient by projecting \eqref{eq:functinal_grad} onto the bases $\{\bm\phi_n\}_{n=0}^{N}$ of the function space:
\begin{equation} \label{eq:function_grad}
    \nabla_{\bm\psi} \mathcal{F} = \langle \bar{\nabla}_{\bm \xi}\mathcal{F},\, \bm{\Phi}\rangle_{w}. 
\end{equation}

\subsubsection{Trajectory smoothness}
\label{sec:smooth}
Similar to \cite{zucker_chomp_2013, schulman_motion_2014},  penalizing the squared joint velocity over $[0,T]$ yields the smooth part: 
\begin{equation*}
    \mathcal{F}_{\mathrm{smooth}}(\bm \psi)
    = \frac{1}{2} \|\dot{\bm \xi} - \dot{\bm{\varphi}} \|_{w}^2 \\
    = \frac{1}{2} \| \dot{\bm\Phi}(t)\,\bm \psi \|_{w}^2. 
\end{equation*}
It yields a precomputable convex quadratic formula: 
\begin{equation} \label{eq:smooth}
    \mathcal{F}_{\mathrm{smooth}}(\bm \psi)=\bm \psi^\top \mathbf Q\,\bm \psi, \,\,\,\,
    \mathbf Q=\frac{1}{2} \| \dot{\bm\Phi}(t) \|_{w}^{2},
\end{equation}
where $\mathbf Q$ depends only on the chosen basis and horizon, and it is well-conditioned for ordinary orthogonal families.

\subsubsection{Obstacle avoidance}
\label{sec:obs} 
\begin{figure}
    \centering
    \begin{subfigure}[t]{0.40\linewidth}
        \includegraphics[width=1\linewidth]{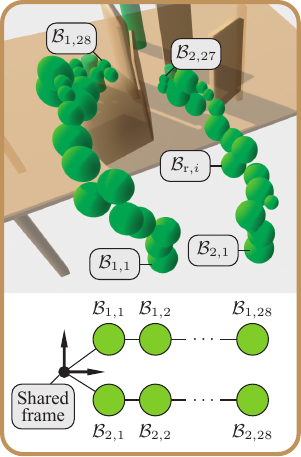}
        \caption{Environment check}
        \label{fig:env_check}
    \end{subfigure}
    \hfill
    \begin{subfigure}[t]{0.58\linewidth}
        \includegraphics[width=1\linewidth]{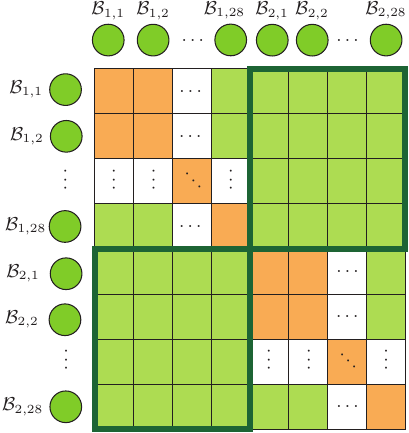}
        \caption{Self check}
        \label{fig:self_check}
    \end{subfigure}
    \caption{
    Environment- and self-collision check for the dual-arm system. 
    (a) Each robot $r$ is represented by its sphere tree $\bm{\mathcal B}_\mathrm{r}$, all rooted in a shared frame. 
    Environment distances are obtained by evaluating each sphere $\mathcal{B}_{\mathrm{r}, i}$ in $\bm{\mathcal B}_\mathrm{r}$ against the workspace SDF. 
    (b) For the self-collision check, all collision-check bodies are vectorized into the unified list 
    $\bm{\mathfrak B} = \bigoplus_{\mathrm{r}=1}^{2} \bm{\mathcal B}_\mathrm{r}$. 
    A multi-robot allowed-collision matrix $A_{cm}$ determines which pairs in $\bm{\mathfrak B}$ must be tested. 
    Green cells denote required distance queries, while orange cells indicate pairs excluded from checking. 
    The structure captures the collision relationships of both intra-robot and cross-robot (shown in the off-diagonal blocks outlined in bold green) within a single framework.
    }
    \label{fig:collision_check}
\end{figure}

This study still uses the sphere tree $\bm{\mathcal{B}} = \{\mathcal{B}\}_{i = 1}^{N_{\mathcal{B}}}$, geometrically consisting of $N_{\mathcal{B}}$ serial collision-check balls (CCB), and the signed distance field (SDF) for single-robot collision checks, as shown in \figref{fig:env_check}.  In this way, let $\bm x(\mathcal B_i,t) \,:  \bm\Theta \rightarrow \bm{\mathcal E}$  be the forward kinematics from the configuration space $\bm\Theta \subset \mathbb{R}^{M}$ to the Euclidean space $\bm{\mathcal{E}} \subset \mathbb{R}^{3}$ where the $i$-th CCB is located. Then, combining the signed distance $d(\bm x)$ with the standard soft cost with buffer $\epsilon$, the collision cost is calculated via the $p$-order piecewise function: 
\begin{equation}
    c(\bm x) = 
    \begin{cases}
        \epsilon - d(\bm x) & d(\bm x) < 0 \\
        \frac{1}{p\,\epsilon^{p-1}}(\epsilon - d(\bm x))^{p} &  0 \leq d(\bm x) \leq \epsilon \\
        0 & \mathrm{otherwise}. 
    \end{cases}
\end{equation}

Different from \cite{zucker_chomp_2013, mukadam_continuous-time_2018}, which combine the collision cost $c(\bm x)$ and the arc length $\int \|\dot{\bm x}\| \diff t$ for the obstacle part, this study generalizes the arc length part  $q\in [0,1]$ to avoid local minima caused by the combination while maintaining the arc length consideration:
\begin{equation} \label{eq:obs0}
    \begin{aligned}
        \mathcal{F}_{\mathrm{obs}}(\bm \psi)
        &= \int_{0}^{T} f_o(\bm\Phi(t) \, \bm\psi) \diff t \\
        &= \int_{0}^{T} \sum_{\mathcal B_i \in \bm{\mathcal B}} c\big(\bm x(\mathcal B_i,t)\big)\ \|\dot{\bm x}(\mathcal B_i,t)\|^{q}\,\mathrm dt, \\
        \dot{\bm x}(\mathcal B_i,t) &=\mathbf J_i \big(\bm\Phi(t) \, \bm\psi \big) \cdot \dot{\bm\Phi}(t) \, \bm\psi, 
    \end{aligned}
\end{equation}
where $\mathbf J_i = \partial_{\bm \theta} \bm x$ is the Jacobian of the $i$-th CCB w.r.t. the joint configuration $\bm \theta = \bm\Phi(t) \, \bm\psi$ at $t$. Then, the obstacle function is calculated only based on the collision cost when $q = 0$, while both the cost and arc length are considered when $q = 1$.  

Since there is still no analytic solution for \eqref{eq:obs0}, this study adopts the discrete-time form and approximates the integral with a quadrature on nodes $\{t_k,w_k\}$: 
\begin{equation} \label{eq:quad_obs}
    \widehat{\mathcal{F}}_{\mathrm{obs}}(\bm \psi)=\sum_{k=1}^{K_\text{obs}} w_k\, f_o \big(\bm\Phi(t_k)\bm \psi\big)^2.  
\end{equation}
Then, the Gauss–Newton linearization and quasi-Hessian estimation can be applied to convexify the objective. 

\subsubsection{Multi-robot collision avoidance} 
\label{sec:multi_robot_obs}

\figref{fig:collision_check} illustrates the collision-checking process for the dual-arm system. 
To maintain computational efficiency, the multi-robot system is decomposed into individual robot instances, each equipped with its own sphere-tree representation. 
For a robot with index $\mathrm{r}$, let \( \bm{\mathcal B}_\mathrm{r} = \{\mathcal B_{\mathrm{r},i}\}_{i=1}^{N_{\mathcal B_\mathrm{r}}} \) denote the set of collision-checking bodies (CCBs) in its sphere tree.  
All sphere trees share the same world-frame root, forming a set of parallel trees  \( \bm{\mathcal{B}}_{1:N_\mathfrak{R}} = \{\bm{\mathcal B}_\mathrm{r}\}_{\mathrm{r}=1}^{N_\mathfrak{R}}\), as shown in \figref{fig:env_check}.  
Environment checking then proceeds by applying the signed-distance evaluation \eqref{eq:obs0} to each robot's sphere tree independently.

To model interactions between robots, the individual sphere trees are stacked into a single vectorized list,
\[
\bm{\mathfrak B} 
= \bigoplus_{\mathrm{r}=1}^{N_\mathfrak{R}} \bm{\mathcal B}_\mathrm{r}
= \bigoplus_{\mathrm{r}=1}^{N_\mathfrak{R}} \bigoplus_{i=1}^{N_{\mathcal B_\mathrm{r}}} \mathcal B_{\mathrm{r},i},
\]
which enumerates all CCBs in the system.  
This allows collision relationships, both within and across robots, to be expressed in a unified binary matrix.

A multi-robot allowed-collision matrix (ACM), denoted $A_{cm}$ and visualized in \figref{fig:self_check}, specifies which CCB pairs must be tested.  
Each entry $A_{cm}(i,j) = \texttt{true}$ (green) indicates that the distance between the $i$-th and $j$-th CCBs in $\bm{\mathfrak B}$ should be evaluated, while $A_{cm}(i,j)=\texttt{false}$ (orange) marks pairs that can be skipped.  
This matrix extends the standard single-robot ACM by including cross-robot and intra-robot collision checks, as well as each robot's internal collision structure.

At each time $t$, all required pairwise distances determined by $A_{cm}$ are computed.  
For a given body $\mathcal B_i$, the self-collision distance is defined as
\[
d_{\mathrm{self}}\!\left( \bm{x}(\mathcal B_i, t) \right)
= \min_{\mathcal B_j \in \{A_{cm}\odot\bm{\mathfrak B}\}\setminus\mathcal B_i}
\mathrm{dist}(\mathcal B_i, \mathcal B_j),
\]
where $\odot$ denotes elementwise filtering using the ACM.  
The resulting distances are cached back into the corresponding robot-level sphere trees $\bm{\mathcal{B}}_{1:N_\mathfrak{R}}$ for use in subsequent evaluations.

Finally, the obstacle cost \eqref{eq:obs0} is computed by combining the environment distance from the SDF $d(\cdot)$ with the self-collision distance $d_{\mathrm{self}}(\cdot)$ obtained from the ACM-filtered sphere-tree checks.  
This provides a consistent mechanism for enforcing both environmental safety and multi-robot collision avoidance during trajectory optimization.

\subsection{Working Objective}
\label{sec:work_obj}
Combining the precomputable quadratic smoothness \eqref{eq:smooth} and the discrete surrogate for obstacles \eqref{eq:obs0}, the objective becomes
\begin{equation} \label{eq:work_obj}
    \widehat{\mathcal F}(\bm \psi)\;=\; \frac{\varrho}{2}\,\bm \psi^\top\mathbf Q\,\bm \psi\;+\;\sum_k w_k\, f_o\!\big(\bm\Phi(t_k)\bm \psi\big)^2. 
\end{equation}
In the multi-robot case, the above objective function is used for a single robot instance based on the decoupled collision-check block (\secref{sec:multi_robot_obs}). 
A batch of objectives for multi-robot trajectory optimization is optimized with Gauss–Newton steps, analytic null-space projections for equality feasibility, and a compact active set for trajectory-wide inequalities. This coefficient-space formulation preserves continuous-time reasoning without dense waypoint sampling.

\section{\facto: Constraint construction}
\label{sec:constraint}

\subsection{Boundary constraints}
\label{sec:bound_constr}
We impose boundary conditions directly in coefficient space using the lifted basis operator $\bm\Phi(t)$ (\secref{sec:coef_space_generic}). With start and goal joint states $\bm\theta_0,\bm\theta_g$, the hard equations are
\[
\bm\Phi(0)\,\bm \psi=\bm\theta_0 - \bm\varphi(0),\qquad
\bm\Phi(T)\,\bm \psi=\bm\theta_g - \bm\varphi(T).
\]
In many tasks, it is advantageous to reduce actuation transients at the boundaries. Let the joint dynamics be $\mathbf M(\bm\xi)\ddot{\bm\xi}+\bm \Psi(\bm\xi,\dot{\bm\xi})\dot{\bm\xi}+\mathbf g(\bm\xi)=\bm u$. A sufficient and simple device is to constrain boundary accelerations and velocities to zero: 
\begin{equation}
    \begin{aligned}
        \ddot{\bm\Phi}(0)\,\bm \psi = \ddot{\bm\Phi}(T)\,\bm \psi=\bm 0 \\
        \dot{\bm\Phi}(0)\,\bm \psi=\dot{\bm\Phi}(T)\,\bm \psi=\bm 0,
    \end{aligned}
\end{equation}
which are linear equalities in $\bm \psi$ and are enforced exactly by null-space projection within each Gauss--Newton step. When a boundary lift is used (Sec.~\ref{sec:orthobase_traj}), the position equalities are satisfied analytically and only the desired boundary derivatives are retained as explicit constraints.

\subsection{Task constraints}
\label{sec:task_constr}
The task constraints from the specific task requirements, such as cup-holding and dual-arm collaboration, are encoded in the compact $\mathrm{SE}(3)$ group \cite{murray_mathematical_2017}. This section first specifies the task requirements and their encoding methods. It then provides evidence of the uniqueness of the locally optimal trajectory and an analysis of task errors.

\subsubsection{Encoding in $\mathrm{SE}(3)$ and $\mathfrak{se}(3)$}
Since all the task constraints for trajectory optimization that this study focuses on are applied to the end-effector, we use the feasible ranges of Euler angles in the ZYX-order, as well as position, to define the end-effector task: 
\begin{equation} \label{eq:task_require}
    \bm{\chi} = [\bm x_{\min}\quad \bm x_{\max}] = 
    \begin{bmatrix}
        \vartheta_z^{\min} & \vartheta_z^{\max} \\
        \vartheta_y^{\min} & \vartheta_y^{\max} \\
        \vartheta_x^{\min} & \vartheta_x^{\max} \\
        p_x^{\min} &  p_y^{\max} \\
        p_y^{\min} &  p_y^{\max} \\
        p_z^{\min} &  p_z^{\max}
    \end{bmatrix}. 
\end{equation}
Then, given the current posture $\bm x = [\vartheta_z, \vartheta_y, \vartheta_x, x, y, z]^\top$, adopting the active set method yields the desired posture 
\begin{equation} \label{eq:desired_posture}
    \bm{x}_\mathrm{des} = \max\big(\min(\bm x, \bm x_{\max}), \, \bm x_{\min} \, \big). 
\end{equation}
It means that all equality and inequality task requirements are generalized as a feasible subspace $\bm\chi$ by introducing the desired posture $\bm{x}_{\mathrm{des}}$, which is activated by the subspace boundary. 

Then, the desired transformation matrix $\mathbf{T}_{\mathrm{des}}$ in $\mathrm{SE}(3)$ and its corresponding twist $\bm \zeta_{\mathrm{des}}$ in $\mathfrak{se}(3)$ can be expressed as
\begin{equation} \label{eq:desired_twist}
    \mathbf{T}_{\mathrm{des}} = 
    \begin{bmatrix}
        \mathbf{R}_{\mathrm{des}} & \mathbf{p}_{\mathrm{des}} \\
        \mathbf{0}^\top & 1
    \end{bmatrix}, 
    \quad
    \bm \zeta_{\mathrm{des}} = \log^{\vee} (\mathbf{T}_{\mathrm{des}}), 
\end{equation}
where $\mathbf{R}_{\mathrm{des}} = \mathbf{R}_z(\vartheta_z^{\mathrm{des}}) \mathbf{R}_y(\vartheta_y^{\mathrm{des}}) \mathbf{R}_x(\vartheta_x^{\mathrm{des}})$, and $\mathbf{R}_z$,  $\mathbf{R}_y$ and $\mathbf{R}_x$ in $\mathrm{SO}(3)$ denote the elemental rotation about axes $z$, $y$ and $x$, respectively, and $\wedge$ and $\vee$ denote the hat and vee operators for gaining the corresponding skew matrix and vector \cite{murray_mathematical_2017}. 

Given the posture $\bm x$, the current transformation matrix $\mathbf{T}$ and its twist $\bm\zeta$ can be calculated in the same way as \eqref{eq:desired_twist}. So, the task constraint, encoded in $\mathrm{SE}(3)$ from \eqref{eq:task_require}, is
\begin{equation} \label{eq:task_constr}
    \bm{h} := \bm{\zeta} - \bm{\zeta}_{\mathrm{des}} = \mathbf 0. 
\end{equation}

In all, the task requirements, generally described by $\bm\chi$, are finally transferred into an equality constraint form $\mathfrak{se}(3)$. It makes the task-constraint expression and its linearization much more concise and compatible with current robotic dynamics libraries, such as Pinocchio \cite{carpentier_pinocchio_2019}. 

\subsubsection{Time-continuous task constraint}
\label{sec:traj_task_constr}
Since \eqref{eq:task_constr} only defines the task constraint given the task requirement $\bm \chi$ \eqref{eq:task_require} for any waypoint, this section expands the task constraints from the waypoint scale to the trajectory scale. 

This study primarily focuses on the trajectory-wide task, which is a stricter requirement. Combining \eqref{eq:traj} and \eqref{eq:task_constr} yields
\begin{equation} \label{eq:task_constr_traj}
    \bm{h}(t) := \big(\bm{\zeta}(\bm\xi) - \bm{\zeta}_{\mathrm{des}}(\bm\xi) \big)(t) = \mathbf 0, \,\,
    \bm\xi(t) = (\bm \varphi + \bm\Phi\bm\psi)(t). 
\end{equation}

Since the function space constructed by $\{\bm{\phi}(t)\}_{n=0}^{N}$ is complete when $N \rightarrow \infty$ due to \secref{sec:coef_space_generic}, it can be concluded that there always exists a solution $\bm \psi$ under the task constraints \eqref{eq:task_constr_traj} as long as the subspace on the manifold constructed by the constraints is contained inside the joint limits. 

However, this study employs a truncated set of basis functions, i.e., $N$ is finite, to mitigate the side effects from the so-called high-frequency noise, cf. \secref{sec:prob_formu}.  Therefore, it is necessary to discretize $\bm{h}(t)$ in \eqref{eq:task_constr_traj} and use a small set of \emph{collocation times} to certify trajectory-wide constraints. 
In this way, $K_\text{tsk}$ discrete nodes $\{t_k\}_{k=1}^{K_\text{tsk}}$  are first collected uniformly over the whole time period $T$, i.e., $t_{k+1} - t_k \equiv {T}/{(K_\text{tsk}+1)}$. Then, they assemble the trajectory-wide task constraints as
\begin{equation}
    \begin{gathered}
        \bm{\mathcal H}(\bm \psi) : = \big[\,\bm h_{1}^\top,\;\dots,\;\bm h_{K_\text{tsk}}^\top \, \big]^\top=\bm 0, \\[6pt]
    \bm h_{k} = \bm h\big(\bm\Phi(t_{k})\bm \psi \big). 
    \end{gathered}
\end{equation}
This set is analytically projected via the null space at each Gauss-Newton step, ensuring feasibility by construction.

\subsection{Joint limit constraints}
\label{sec:limit_constr}
Since there is either no analytic solution for $\|\bm \xi(t)\|_{\infty} = \|\bm\Phi(t) \bm\psi\|_{\infty}$ and $\|\bm \xi(t)\|_{\infty^-} = \|\bm\Phi(t) \bm\psi\|_{\infty^-}$,  this study also adopts the discrete checking form of \cite{feng_optimal_2023} to ensure the almost trajectory-wide joint-limit satisfaction: 
\begin{equation}
    \bm\theta_{\min}^{\oplus K_{\text{lmt}}} \leq  [\bm\Phi(t_1)^\top \dots  \bm\Phi(t_{K_\text{lmt}})^\top]^\top \bm\psi \leq \bm\theta_{\max}^{\oplus K_{\text{lmt}}}, 
\end{equation}
where $t_k = k\cdot T / (K_\text{lmt} + 1)$ and $K_\text{lmt}$ denotes the total amount of sampled waypoints for the joint-limit check. 

We use a small set of check nodes, initialized uniformly and refined adaptively near potential violations, to achieve trajectory-wide feasibility in practice while avoiding dense uniform sampling. To prevent excessive fluctuations from frequent node changes, we update the node set conservatively (only when violations are detected), which keeps the optimization smooth and reliable.

\subsection{Uniqueness of the local optimum under the constraints}
\label{sec:opt_constr}

\begin{figure}
    \centering
    \includegraphics[width=0.95\linewidth]{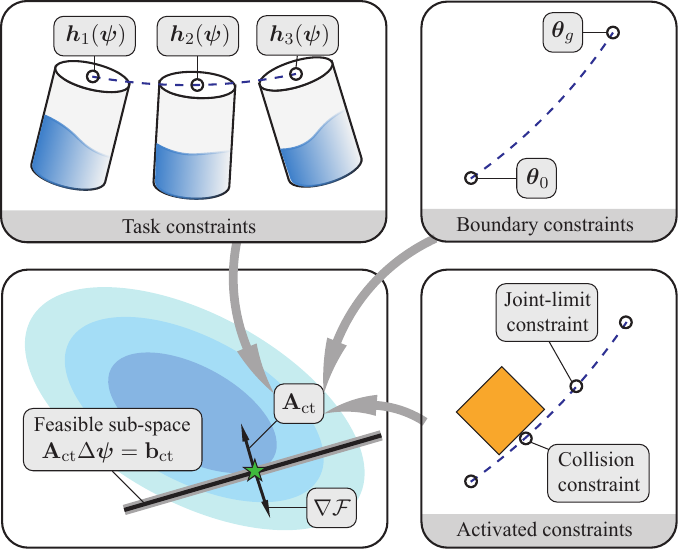}
    \caption{
    Illustration of the constraint structure used in trajectory optimization. 
    Task constraints $h_i(\psi)$ (top left) restrict the solution to trajectories that satisfy task-specific requirements.  
    Boundary constraints (top right) enforce initial and goal conditions on the trajectory coefficients.  
    Activated constraints (bottom right), such as joint limits or collision constraints, become active only when the current iterate approaches the corresponding boundary.  
    All active constraints are collected into the linearized system 
    $A_{\mathrm{ct}} \Delta\psi = b_{\mathrm{ct}}$ (bottom left), which defines a feasible update subspace intersecting with the descent direction of the objective $\nabla \mathcal{F}$. 
    The optimization then proceeds within this subspace to ensure constraint satisfaction while reducing the objective cost.
    }
    \label{fig:constr_opt}
\end{figure}

Given a truncated function space $\bm\Xi' = \mathrm{span}\{\bm\phi\}_{n = 1}^{N}$, let $\bm\xi^\ast  \in \bm\Xi'$ with $\bm\psi^\ast \in \mathbb{R}^{M(N+1)}$ be a set of optimum. 
Given the boundary conditions (\secref{sec:bound_constr}), joint limits (\secref{sec:limit_constr}), and task requirements (\secref{sec:traj_task_constr}), let $K_{\mathrm{bnd}}^*$, $K_{\mathrm{lmt}}^*$, and $K_{\mathrm{tsk}}^*$ be the number of active constraints for the boundary, limits, and task, respectively. 

This section only reserves the smooth part of \eqref{eq:work_obj} and treats the obstacle part as a set of inequality constraints, $\{d(\bm{x}_{i,t}) > \epsilon \}_{i = 1, \, t=0}^{N_{\mathcal{B}}, \, T}$  \eqref{eq:obs0}, because the optimum is collision-free. Then, let $K_{\mathrm{obs}}$ be the number of active obstacle constraints. 

Let $\bm\psi$ be in the convex neighborhood of the optimum $\bm\psi^\ast$ and $K^* = K_{\mathrm{bnd}}^* + K_{\mathrm{lmt}}^* + K_{\mathrm{tsk}}^* + K_{\mathrm{obs}}^*$ be the total number of active constraints. The constrained optimization problem, shown in \figref{fig:constr_opt}, can then be expressed as 
\begin{equation}
    \begin{aligned}
        \Delta\bm{\psi}^\ast =& \argmin_{\Delta\bm\psi} \, \frac{1}{2}(\bm\psi + \Delta \bm\psi)^\top \mathbf Q (\bm\psi + \Delta \bm\psi), \\
        & \mathrm{s.t.} \,\, \mathbf{A}_{\mathrm{ct}} \Delta\bm\psi = \mathbf{b}_{\mathrm{act}}.
    \end{aligned}
\end{equation}
where $\mathbf{A}_{\mathrm{ct}} \in \mathbb{R}^{K^* \times M(N+1)}$ and $\mathbf{b}_{\mathrm{act}} \in \mathbb{R}^{K^*}$ denote the active constraint, linearized at  $\bm{\psi}$. Under the KKT \cite{nocedal_numerical_2006} condition, the optimal $\Delta\bm\psi^\ast$ must satisfy: 
\begin{equation}
    \begin{bmatrix}
        \mathbf{Q} & -\mathbf{A}_{\mathrm{ct}}^\top \\
        \mathbf{A}_{\mathrm{ct}} & \mathbf{O}
    \end{bmatrix}
    \begin{bmatrix}
        \Delta\bm\psi^\ast \\
        \bm \lambda^\ast
    \end{bmatrix}
    = 
    \begin{bmatrix}
        -\mathbf{Q} \bm\psi \\
        \mathbf{b}_{\mathrm{act}} 
    \end{bmatrix}
\end{equation}
where the Lagrangian multiplier $\bm\lambda^\ast \in \mathbb{R}^{+, K_{\mathrm{act}}}$, because all the constraints are active. 
Therefore, $[\Delta\bm\psi^\ast, \, \bm \lambda^\ast]$ exists uniquely as long as $M(N+1) \geq K_{\mathrm{act}}$. It means that the task error and the number of task constraints are determined given a specific planning scenario,  and that the error goes to $0$ when $N \rightarrow \infty$.

\section{\facto: Constrained Trajectory Optimization \\ for Multi-robot}
\label{sec:facto}
The above section introduces how to formulate the objective function in a coefficient space that accounts for both trajectory smoothness and safety, and how to construct constraints for boundary conditions, joint limits, and task requirements. Based on this, the section primarily focuses on \facto (Algorithms \ref{alg:facto} \& \ref{alg:update}), a sequential quadratic programming algorithm in the null space for multi-robot, and introduces an intuitive method for estimating the execution time of the generated motion. 

\begin{algorithm}[!htp]
\caption{FACTO}
\label{alg:facto}
\DontPrintSemicolon
\LinesNumbered
\SetKwInOut{Input}{Input}
\SetKwInOut{Output}{Output}
\SetKw{Break}{break}
\SetKw{And}{and}
\SetKw{Or}{or}
\linespread{1.0}

\Input{Start/goal $\bm\theta_{1:N_\mathfrak{R}}^{0}, \bm\theta_{1:N_\mathfrak{R}}^{g}$; basis $\{\bm\phi(t)\}_{n=0}^{N}$ and lift $\bm\varphi(t)$; obstacle and task nodes $\{t_j,w_j\}$. }
\Output{Coefficients $\bm \psi$.}

\textbf{Init: }
\For{$\mathfrak{R} \in \bm{\mathfrak{R}}$}{
    Extract start/goal: $\bm\theta_0, \bm\theta_g  \leftarrow \bm\theta_{1:N_\mathfrak{R}}^{0}(\mathfrak{R}), \bm\theta_{1:N_\mathfrak{R}}^{g}(\mathfrak{R})$; \;
    Extract smoothness term: $\mathbf Q \leftarrow \mathbf Q_{1:N_\mathfrak{R}}(\mathfrak{R})$; \;
    \(
    \begin{aligned}
        \bm \psi_0 &= \mathop{\argmin} \nolimits_{\bm \psi} \,\bm \psi^\top\mathbf Q\bm \psi, \, \\[-3pt]
        & \mathrm{s.t.}\, \bm\Phi(0)\bm \psi=\bm\theta_0,\ \bm\Phi(T)\bm \psi=\bm\theta_g; 
    \end{aligned}
    \) \\[2pt] 
    Set $\bm\psi_{1:N_\mathfrak{R}}(\mathfrak{R}) = \bm \psi_0$, $\overline{\mathbf{H}}\!=\!\eta \mathbf{I}$, $\overline{\bm g}\!=\!\bm 0$, $\lambda\!=\!\lambda_0$.
}

\For(\tcp*[h]{Sequential QP}){$i=0,1,\dots$ }{
    Multi-robot collision check and update the collision information of $\bm{\mathcal{B}}_{1:N_\mathfrak{R}}$ (\secref{sec:multi_robot_obs}); \;
    Calculate task constraints $\bm{\mathcal{H}}_{1:N_\mathfrak{R}}$; \;
    \For(\tcp*[h]{For each robot}){$\mathfrak{R} \in \bm{\mathfrak{R}}$ }{
        Extract coefficient: $\bm\psi \leftarrow \bm\psi_{1:N_\mathfrak{R}}(\mathfrak{R})$; \;
        $\bm\psi, \Delta\bm\psi$ = \code{Update($\bm\psi$,$\bm{\mathcal{B}}_{1:N_\mathfrak{R}}$,$\bm{\mathcal{H}}_{1:N_\mathfrak{R}}$,$\mathfrak{R}$)}; \;
        Update the coefficient and step by $\bm \psi_{1:N_\mathfrak{R}}(\mathfrak{R}) \leftarrow \bm{\psi}$, $\Delta\bm \psi_{1:N_\mathfrak{R}}(\mathfrak{R}) \leftarrow \Delta\bm{\psi}$; \;
    }
    \If{$\|\Delta\bm \psi_{1:N_\mathfrak{R}} \|\le\texttt{stepTol}$ \Or stationarity met}{
        \Break.  \;
    }
}
Estimate the execution time by $T = \code{TimeScaler($\bm{\psi}_{1:N_\mathfrak{R}}$,$\bm{V}^{\lim}$,$\bm{U}^{\lim}$,$\gamma$)}$. 
\end{algorithm}
\begin{algorithm}[!htp]
\caption{Update}
\label{alg:update}
\DontPrintSemicolon
\LinesNumbered
\SetKwInOut{Input}{Input}
\SetKwInOut{Output}{Output}
\linespread{1.0}

\Input{Coefficient $\bm\psi$, sphere trees $\bm{\mathcal{B}}_{1:N_\mathfrak{R}}$, task constraints $\bm{\mathcal{H}}_{1:N_\mathfrak{R}}$, robot instance $\mathfrak{R}$. }
\Output{Updated coefficient $\bm \psi$, update step $\Delta\bm \psi$.}

Extract info: $\bm{\mathcal{B}} \leftarrow \bm{\mathcal{B}}_{1:N_\mathfrak{R}}(\mathfrak{R})$, $\bm{\mathcal{H}} \leftarrow \bm{\mathcal{H}}_{1:N_\mathfrak{R}}(\mathfrak{R})$; \;
\tcp{1) Obstacle residuals}
\For {$\{t,w\} \in \{t_k,w_k\}_{k=1}^{K_\text{obs}}$}{
Calculate the residual $r_j$ from the sphere tree $\bm{\mathcal{B}}$  and do the linearization by \eqref{eq:lin_res}; \;
Estimate the gradient $\widehat{\mathbf g}$ and Hessian $\widehat{\mathbf H}$ of the obstacle function by \eqref{eq:obs_grad_Hessian}; \;
}

\tcp{2) EMA smooth}
Calculate the smoothed gradient $\overline{\mathbf g}$ and Hessian $\overline{\mathbf H}$ of the obstacle function by \eqref{eq:ma}; \;

\tcp{3) Linearize equalities and build reduced coordinates}
Stack all the linearized equalities (boundary and active task constraints by \eqref{eq:constr_stack} and gain $\mathbf A_{\mathrm{eq}}\Delta\bm \psi=\bm b_{\mathrm{eq}}$.\;
Get a feasible step $\Delta\bm \psi_0$ and the null space $\mathbf N$ of the equations $\mathbf A_{\mathrm{eq}}$; \;

\tcp{4) Reduced GN QP in $\bm z$}
Gain the reduced gradient $\mathbf g_{\mathrm{red}}$ and Hessian $\mathbf H_{\mathrm{red}}$ of the objective by \eqref{eq:model_red}; \;

\tcp{5) Project inequalities}
Form the inequalities $\mathbf A_{\mathrm{ieq}} \Delta\bm \psi\le\mathbf{b}_{\mathrm{ieq}}$ and gain the reduced inequalities $\mathbf A_{\mathrm{ieq}}^{\mathrm{red}} \bm{z} \leq \mathbf{b}_{\mathrm{ieq}}^{\mathrm{red}}$ by \eqref{eq:ieq_red}; \; 

\tcp{6) Solve reduced QP for $r$-th robot}
$
\begin{aligned}
  \bm z^\star\!\leftarrow\!\arg\min_{\bm z}\ & \tfrac{1}{2}\bm z^\top\mathbf H_{\mathrm{red}}\bm z + \bm g_{\mathrm{red}}^\top\bm z, \\
  \mathrm{s.t.}\ & \mathbf A_{\mathrm{ieq}}^{\mathrm{red}}\bm z\le\bm h_{\mathrm{red}}; 
\end{aligned}
$  \;
Compose step $\Delta\bm \psi\!\leftarrow\!\Delta\bm \psi_0+\mathbf N\bm z^\star$; \;

\tcp{7) Accept and update}
Estimate the trust ratio $\rho\!=\!{\textit{ared}}/{\textit{pred}}$ by \eqref{eq:ratio}; \;
\lIf{$\rho\ge\rho_{\max}$}
{ $\bm \psi\!\leftarrow\!\bm \psi+\Delta\bm \psi$;\ $\lambda\!\leftarrow\!\gamma_{\downarrow}\lambda$;}
\lElseIf{$\rho\le\rho_{\min}$}
{ $\lambda\!\leftarrow\!\gamma_{\uparrow}\lambda$. }

\end{algorithm}
%

\subsection{Convexification of Objective Function}
\label{sec:ema-gn}
Though \secref{sec:work_obj} introduces the work objective \eqref{eq:work_obj} for trajectory optimization in a coefficient space, it is still hard to find its optimum numerically due to the non-convexity. Therefore, we minimize the objective using a Gauss-Newton (GN). The obstacle term is written as a sum of squared residuals on the selected time points (\secref{sec:obs}), while the smoothness term contributes a known convex quadratic. 

\subsubsection{Residual linearization}
Let $\{t_k,w_k\}_{k=1}^{K_\text{obs}}$ be time points and weights, and $r_k(\bm \psi)=\sqrt{w_k}\,f_o\!\big(\bm\Phi(t_k)\bm \psi\big)$ be the scalar residual at node $k$ with $\bm\Phi(t_k)$. 
Then, the linearized model is 
\begin{equation} \label{eq:lin_res}
    \begin{aligned}
        r_k(\bm \psi, \Delta\bm \psi)
        &\;\approx\; r_k(\bm \psi) + \bm{g}_k^\top \,\Delta\bm \psi \\
        &= r_k(\bm \psi) + \Big(\partial_{\bm\theta} f_o\,\Big|_{\bm\theta=\bm\Phi(t_k)\bm \psi}\Big)^\top
        \,\bm\Phi(t_k)\,\Delta\bm \psi,
    \end{aligned}
\end{equation}

\subsubsection{Gauss-Newton term}
Based on the quadratic form of \eqref{eq:quad_obs},  the residual linearization constructs a GN term at an optimization step: 
\begin{equation*}
    \begin{aligned}
        \widehat{\mathcal{F}}_{\mathrm{obs}}(\bm \psi)
        &= \sum_{k} \big( r_k(\bm\psi) + \mathbf{g}_k^\top \Delta \bm\psi \big)^2 + \mathcal{O}(\|\Delta \bm\psi \|^4)
    \end{aligned}
\end{equation*}
Then, the GN standard components are
\begin{equation} \label{eq:obs_grad_Hessian}
    \widehat{\mathbf H}=\sum_{k} \mathbf{g}_k \mathbf{g}_k^\top \succeq 0,\qquad
    \widehat{\mathbf g}=\sum_{k} r_k \, \mathbf{g}_k.
\end{equation}
These capture the curvature and gradient of $\bm \psi$.

\subsubsection{Exponential moving average (EMA)}
To reduce the noise caused by frequent switches between the active (in-collision) and inactive (collision-free) CCB statuses during the GN term estimation, we adopt the unbiased EMA \cite{feng_optimal_2023} to smooth the sequential estimation of the Hessian matrix and gradient: 
\begin{equation} \label{eq:ma}
    \overline{\mathbf g} = \frac{\sum_{\iota = 1}^{i} \beta_1^{\,\iota} \cdot \widehat{\mathbf g}_\iota}{1 - (1-\beta_1)^i},
    \quad
    \overline{\mathbf H} = \frac{\sum_{\iota = 1}^{i} \beta_2^{\,\iota} \cdot \widehat{\mathbf H}_\iota}{1 - (1-\beta_2)^i}
\end{equation}
with $\beta_1, \beta_2 \in(0,1)$.

According to the smoothness function of \secref{sec:smooth}, its gradient and constant PSD curvature are $\varrho \mathbf{Q} \bm\psi$  and $\varrho\mathbf{Q}$, respectively. With a small LM-style damping $\lambda\ge 0$, the convex surrogate around $\bm \psi$ reads
\begin{equation}
    \begin{aligned}
        m(\Delta\bm \psi)
        &= \tfrac{1}{2} \Delta\bm \psi^\top\! (\varrho \mathbf Q+ \overline{\mathbf H} + \lambda \mathbf I) \Delta\bm \psi \\
        &+ (\varrho \mathbf{Q} \bm\psi+ \overline{\mathbf g})^\top \Delta\bm \psi. 
    \end{aligned}
\end{equation}

\subsection{Null-space Trajectory Optimization}

The above section explains how to approximate the nonlinear objective function with a smooth and convex function. This section mainly focuses on how the null space method is applied to the approximated function.

\subsubsection{Equality constraints and reduced coordinates}
In this study, there are two types of equality constraints: boundary constraints (\secref{sec:bound_constr}) and task constraints (\secref{sec:task_constr}). According to \secref{sec:opt_constr}, the number of boundary constraints and the selected task constraints is determined to get a unique optimum for the whole constrained optimization problem. Therefore, considering the nonlinearity of the task error $\bm{h}(\bm\psi)$ in \eqref{eq:task_constr}, we first linearize it at $\bm\psi$ by 
\begin{equation} \label{eq:lin_task_constr}
    \bm h(\bm \psi + \Delta\bm \psi)
    \;\approx\;\bm h(\bm \psi) \;+\; \underbrace{\Big(\partial_{\bm\theta}\bm h \Big)^\top}_{\text{task Jacobian}}\,\bm\Phi(t)\,\Delta\bm \psi 
    \;=\;\bm 0, 
\end{equation}
where $\bm\theta=\bm\Phi(t)\bm \psi$ and $\partial_{\bm\theta}\bm h$ is the Jacobian of the task error in $\mathfrak{se}(3)$ \cite{murray_mathematical_2017}. Then, the matrix $\mathbf{A}_{\mathrm{eq}}$ and vector $\mathbf{b}_{\mathrm{eq}}$ stack the boundary and linearized task constraints:
\begin{equation} \label{eq:constr_stack}
    \begin{gathered}
        \mathbf A_{\mathrm{eq}}\,\Delta\bm \psi\;=\;\mathbf{b}_{\mathrm{eq}},\\[6pt]
        \mathbf A_{\mathrm{eq}} =
        \begin{bmatrix}
            [\bm\Phi^\top(0), \, \bm\Phi^\top(T)]^\top \\
            [\dot{\bm\Phi}^\top(0), \, \dot{\bm\Phi}^\top(T)]^\top \\
            \partial_{\bm\theta}\bm h_1\,\bm\Phi(t_{1})\\
            \vdots\\ 
            \partial_{\bm\theta}\bm h_{K'}\,\bm\Phi(t_{K_\text{tsk}})
        \end{bmatrix},
        \,
        \mathbf{b}_{\mathrm{eq}} =
        -\begin{bmatrix}
            \bm 0 \\
            \bm 0 \\
            \bm h_1(\bm \psi)\\ 
            \vdots\\ 
            \bm h_{K_\text{tsk}}(\bm \psi)
        \end{bmatrix}.
    \end{gathered}
\end{equation}

\begin{figure}[!htp]
    \centering
    \includegraphics[width=0.95\linewidth]{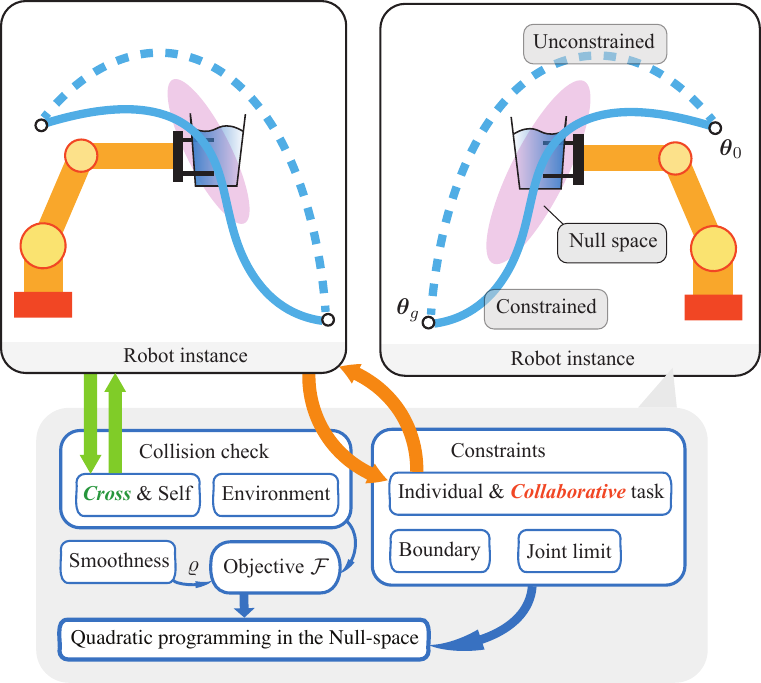}
    \caption{
    Overview of the \facto optimization framework used for multi-robot trajectory generation. 
    Each robot instance performs collision checking (environment, self, and cross-robot) and enforces individual or collaborative task constraints together with boundary and joint-limit conditions. 
    These components contribute to the objective $\mathcal{F}$ and the associated smoothness term $\varrho$. 
    Active constraints are handled through a null-space projection, after which a quadratic program is solved in the remaining feasible subspace. 
    The lower panel illustrates the resulting motion update: the unconstrained search direction (blue dash) is projected onto the null space to satisfy active constraints, yielding the constrained update (blue) between the initial configuration $\bm q_0$ and the goal $\bm q_g$.
    }
    \label{fig:null_space}
\end{figure}

Then, we compute (i) a minimum-norm \emph{feasibility step} $\Delta\bm \psi_0=\mathbf A_{\mathrm{eq}}^\dagger \mathbf b_{\mathrm{eq}}$ and (ii) a \emph{null-space basis} $\mathbf N$ generated by the QR decomposition \cite{nocedal_numerical_2006, noauthor_eigen_nodate} such that $\mathrm{Null}(\mathbf A_{\mathrm{eq}})=\mathrm{range}(\mathbf N)$. Any feasible step decomposes as
\[
\Delta\bm \psi \;=\; \Delta\bm \psi_0 \;+\; \mathbf N\,\bm z,
\]
so the search is reduced to the smaller vector $\bm z$ whose dimension equals the nullity of $\mathbf A_{\mathrm{eq}}$.
As shown in \figref{fig:null_space}, this projection enforces all equalities \emph{exactly} at every GN iteration and shrinks the feasible searching space for trajectory optimization, especially under task constraints.

\subsubsection{Reduced Gauss--Newton model with EMA curvature}
\secref{sec:ema-gn} provides a convex GN surrogate at $\bm \psi$ with a precomputable smoothness quadratic $\mathbf Q$, the EMA-smoothed GN terms $\overline{\mathbf H},\overline{\bm g}$  from obstacle residuals, and a small LM damping $\lambda$. Substituting $\Delta\bm \psi=\Delta\bm \psi_0+\mathbf N\bm z$ yields a reduced, strictly convex model in $\bm z$: 
\begin{equation} \label{eq:model_red}
    \begin{aligned}
        m^{\mathrm{red}}(\bm z) &\;=\;\tfrac{1}{2}\bm z^\top \underbrace{\mathbf N^\top(\mathbf Q+\overline{\mathbf H}_k+\lambda \mathbf I)\mathbf N}_{\mathbf H_k^{\mathrm{red}}}\bm z \\
        & \;+\;\underbrace{\Big(\mathbf N^\top(\mathbf Q+\overline{\mathbf H}+\lambda \mathbf I)\Delta\bm \psi_0+\mathbf N^\top(\bm s+\overline{\bm g})\Big)}_{\bm g^{\mathrm{red}}}{}^\top\bm z \\
    \end{aligned}
\end{equation}
All matrices are small (nullity-sized) and inherit the conditioning benefit of EMA smoothing.

\subsubsection{Inequalities in the reduced space}
Trajectory-wide \emph{inequality} constraints are imposed at the same collocation set: 
joint envelopes. Each linear/SOCP constraint in $\Delta\bm \psi$ is projected to $\bm z$ by left-multiplication with $\mathbf N$ and a shift by $\Delta\bm \psi_0$: 
\begin{equation} \label{eq:ieq_red}
    \mathbf A_{\mathrm{ieq}}\,\Delta\bm \psi \le \bm b_{\mathrm{ieq}} \,
    \Rightarrow\, 
    \mathbf A_{\mathrm{ieq}} \big(\Delta\bm \psi_0+\mathbf N\bm z\big)\le \mathbf b_{\mathrm{ieq}},
\end{equation}
which remains linear in $\bm z$. Because the envelopes act in coefficient space, feasibility can be efficiently enforced using a small (adaptive) set of critical time nodes, avoiding dense, uniform sampling while maintaining trajectory-wide constraint satisfaction in practice.

\subsubsection{Operator-splitting step and globalization}
\facto (\algref{alg:update}) minimizes $m^{\mathrm{red}}(\bm z)$ in \eqref{eq:model_red} subject to the projected inequalities using the warm-started ADMM \cite{banjac_infeasibility_2019} and OSQP \cite{stellato_osqp_2020}, an operator-splitting solver. A trust-region ratio 
\begin{equation} \label{eq:ratio}
    \rho 
    = \frac{\textit{ared}}{\textit{mred}}
    = \frac{\mathcal{F}(\bm\psi + \Delta \bm\psi_0 + \mathbf{N}\bm z) - \mathcal{F}(\bm\psi)}{m^\mathrm{red}(\bm z)}
\end{equation}
controls the damping $\lambda$:  poor agreement increases $\lambda$ to emphasize stability; good agreement decreases $\lambda$. Accepted steps update $\bm \psi \leftarrow \bm \psi + \Delta\bm \psi$. 

\subsubsection{Outcome}
The combination of analytic null-space projection (exact equality feasibility), coefficient-space envelopes (time-continuous parameterization with sparse enforcement), and EMA-smoothed GN curvature (robust convergence) yields fast, well-conditioned steps that refine the \emph{entire} trajectory with a handful of coefficients, even in the presence of task constraints in $\mathrm{SE}(3)$.

\subsection{Multi-robot consideration}

The above sections introduce how \facto deals with the constrained nonlinear trajectory optimization problem of a single-robot system. In this section, we treat the multi-robot system $\bm{\mathfrak{R}}$ as a collection of decoupled single-robot systems $\{\mathfrak{R}_\mathrm{r}\}_{\mathrm{r} = 1}^{N_\mathfrak{R}}$, each treated as a robot instance. Each instance $\mathfrak{R}$ consists of a sphere tree $\bm{\mathcal{B}}$ for collision check, an objective $\mathcal{F}(\bm\psi)$ \eqref{eq:obj0} affiliated with its work objective $\widehat{F}(\bm\psi)$ \eqref{eq:work_obj}, boundary constraints (\secref{sec:bound_constr}), joint-limit constraints (\secref{sec:limit_constr}), and task constraints $\bm{\mathcal{H}}(\bm\psi)$ (\secref{sec:task_constr}). As shown in \figref{fig:null_space}, different instances share their sphere tree information for cross-robot self-collision check and their end-effector postures for the collaborative task. All the data, calculated from the shared information, is then cached back to each instance. Based on that, the trajectory coefficient $\bm\psi$ of each instance $\mathfrak{R}$ is optimized, which updates the trajectory bundle $\bm\psi_{1:N_\mathfrak{R}}$, until the iteration meets the stationarity condition $\frac{\|\Delta\bm\psi_{1:N_\mathfrak{R}}\|}{\bm\psi_{1:N_\mathfrak{R}}} \leq \code{stepTol}$.

\subsection{Execution Time via Sampled Maximum–Torque Scaling}
\label{sec:time-scaling-sampled}
Though the optimal trajectory $\bm\xi = \bm\Phi(t) \bm\psi$ is a continuous function over $[0,T]$, it still cannot be called an optimal trajectory in a strict sense. That is because the execution time $T$ is not determined. This section introduces the way to adjust the execution time $T$ to meet the joint torque requirement, while keeping the \emph{trajectory shape}, i.e., the coefficients $\bm \psi$,  fixed. The idea of \algref{alg:time-scaling} is simple:  (i) sample torques over $[0,T]$, (ii) estimate the worst (maximum) torque across samples and joints, and (iii) rescale $T$ using the known $1/T^2$ law of inertial/Coriolis terms. 

\begin{algorithm}[!htp]
\caption{TimeScaler}
\label{alg:time-scaling}
\DontPrintSemicolon
\SetKw{Init}{Init: }
\KwIn{Coefficients bundle $\bm \psi_{1:N_\mathfrak{R}}$, velocity and torque limits $\bm V^{\lim}_j, \bm U^{\lim}_j$, safety factor $\gamma\!\in(0,1)$}. 
\KwOut{Adjusted $T$}
\For(\tcp*[h]{For each robot instance}){$\mathfrak{R} \in \bm{\mathfrak{R}}$}{
    Extract coefficient $\bm{\psi} \leftarrow \bm{\psi}_{1:N_\mathfrak{R}}(\mathfrak{R})$; \;
    Extract velocity and torque limits of each joint-$j$: $V^{\lim}_j, U^{\lim}_j \leftarrow \bm V^{\lim}_j(\mathfrak{R}), \bm U^{\lim}_j(\mathfrak{R})$; \;
    $V_j \leftarrow (\dot{\bm{\Phi}} \cdot \bm{\psi} + \dot{\bm\varphi})_j$, $T \leftarrow \max_j (V^{\lim}_j / \gamma V_j)$; \;
    $T_{1:N_\mathfrak{R}}(\mathfrak{R}) \leftarrow T$
}
\Init $T \leftarrow \max(T_{1:N_\mathfrak{R}})$; \;
\For(\tcp*[h]{For each robot instance}){$\mathfrak{R} \in \bm{\mathfrak{R}}$}{
    \For{$k=1,\dots,K_{\max}$}{
      Sample $t_\ell=\ell T/L$, compute $\bm u_\ell$, $\mathbf g(\bm\theta_\ell)$; \;
      $U^{\mathrm{dyn,obs}}_j \leftarrow \max_\ell |u_{\ell,j}-g_{\ell,j}|$,\quad
      $G^{\max}_j \leftarrow \max_\ell |g_{\ell,j}|$,\quad
      $\sigma \leftarrow \max\!\big\{1,\ \sqrt{\max_j U^{\mathrm{dyn,obs}}_j/(\gamma U^{\mathrm{lim}}_j)}\big\}$; \;
      \lIf{$\sigma \le 1$}{\textbf{break};}
      $T \leftarrow \min(\sigma_{\max}\,T,\ \sigma T)$, $T_{1:N_\mathfrak{R}}(\mathfrak{R}) \leftarrow T$; 
    }
}
\Return $\max(T_{1:N_\mathfrak{R}})$
\end{algorithm}

\subsubsection{Initialization}
The joint limit constraints of \secref{sec:limit_constr} ensure all the generated trajectories $\bm{\psi}_{1:N_\mathfrak{R}}$ are in the feasible joint space. However, this study does not consider the joint velocity limit when optimizing $\bm{\psi}_{1:N_\mathfrak{R}}$ due to $T$'s scalability.  Therefore, \code{TimeScaler} initializes $T$ according to the velocity limits of each joint $V_{j}^{\lim}$ and caches them in a vector $T_{1:N_\mathfrak{R}}$. To ensure that all robots can execute within their velocity limits, the initialization adopts the maximum value, i.e., $\max(T_{1:N_\mathfrak{R}})$. 

\subsubsection{Sampling}
Choose $L$ uniform sample times $t_\ell=\ell\,T/L$, $\ell=[0\dots L]$, and evaluate
\[
\bm\theta_\ell=\bm\Phi(t_\ell)\bm \psi,\quad
\dot{\bm\theta}_\ell=\dot{\bm\Phi}(t_\ell)\bm \psi,\quad
\ddot{\bm\theta}_\ell=\ddot{\bm\Phi}(t_\ell)\bm \psi.
\]
Compute the torques at current $t_\ell$ by the inverse-dynamics: 
\begin{equation}
    \bm u_\ell
    =\bm M(\bm\theta_\ell)\,\ddot{\bm\theta}_\ell
    +\bm{C}(\bm\theta_\ell,\dot{\bm\theta}_\ell)\,\dot{\bm\theta}_\ell
    +\bm g(\bm\theta_\ell)
\end{equation}
with the mass matrix $\bm M$, Coriolis $\bm C$, and gravity vector $\bm g$. 

\subsubsection{Separate gravity from dynamic effort}
Since the trajectory shape $\bm\psi$ is constant, we can say that $\bm\theta_\ell$, $\bm M(\bm\theta_\ell)$ and $\bm g(\bm\theta_\ell)$ are also constant, and that $\dot{\bm\theta}_\ell\!\propto\!1/T$ and $\ddot{\bm\theta}_\ell\!\propto\!1/T^2$ because $\dot{\bm\Phi}$ and $\ddot{\bm\Phi}$ already include the time derivatives. So, the extracted dynamic part becomes
\begin{equation}
    \bm u^{\mathrm{dyn}}_\ell \;: =\; \bm u_\ell - \mathbf g(\bm\theta_\ell)
\quad (\propto 1/T^2).
\end{equation}

\subsubsection{Worst-case torque and dynamic margin}
For a joint $j$, 
\[
U^{\mathrm{dyn,obs}}_j \;=\; \max_\ell \big|[\,\bm u^{\mathrm{dyn}}_\ell\,]_j\big|,
\qquad
G^{\max}_j \;=\; \max_\ell \big|[\mathbf g(\bm\theta_\ell)]_j\big|.
\]
Given symmetric joint limits $|u_j|\le U^{\lim}_j$, the maximum dynamic margin allowed is $U^{\mathrm{dyn,lim}}_j : = \varsigma (U^{\lim}_j - G^{\max}_j)$ with a scalar $\varsigma \in (0,1)$ for the execution safety. 

\subsubsection{Closed-form time update}
Because $U^{\mathrm{dyn}} \propto 1/T^2$ the ratio between observed and allowed dynamic effort sets the
required scale factor. With a safety margin $\gamma\in(0,1)$,
\begin{gather}
    \sigma = \max\!\left\{\,1,\ \sqrt{\ \max_{j \in \{1 \dots M\}} \frac{U^{\mathrm{dyn,obs}}_j}{\gamma\,U^{\mathrm{dyn,lim}}_j}\ }\ \right\}, \\[6pt]
    T_{\text{new}} \;=\; \sigma\,T_{\text{old}}.
\end{gather}
If the current trajectory already satisfies all limits, $\sigma=1$ and $T$ are unchanged.

\section{Implementation of \facto}

\subsection{Orthogonal Basis Functions}
\label{sec:facto_basis}

\facto operates in a function space by parameterizing each joint trajectory as a linear combination of time-continuous orthogonal basis functions over $[0,T]$, with a low-order boundary lift enforcing start- and goal-waypoint conditions. In the implementation, three families of orthogonal bases are supported: sine, cosine, and Chebyshev polynomials. All share the common form in \eqref{eq:traj}–\eqref{eq:traj_acc}, where $\bm\phi(t)\in\mathbb{R}^{N+1}$ and its derivatives are evaluated analytically.

\subsubsection{Cosine basis}
\label{sec:cos}
The cosine basis is defined by
\begin{equation}
\phi_n(t) = \cos\!\left(n\pi t/T\right), \quad n=0,\dots,N ,
\end{equation}
which is orthogonal on $[0,T]$ under the standard $L^2$ inner product. The first and second derivatives follow directly as
\begin{align*}
\dot{\phi}_n(t) = -\tfrac{n\pi}{T}\sin\!\left(\tfrac{n\pi t}{T}\right), \quad
\ddot{\phi}_n(t) = -\left(\tfrac{n\pi}{T}\right)^2\cos\!\left(\tfrac{n\pi t}{T}\right).
\end{align*}
This basis naturally represents smooth, non-periodic motions and is used with a linear boundary lift $\varphi(t)$ to satisfy joint position and velocity constraints at $t=0$ and $t=T$.

\subsubsection{Sine basis}
\label{sec:sin}
The sine basis employs
\begin{equation}
\phi_n(t) = \sin\!\left(n\pi t/T\right), \qquad n=1,\dots,N+1 ,
\end{equation}
which vanishes at both endpoints. When combined with the boundary lift, the coefficients $\bm\psi$ describe only the interior deformation of the trajectory, while endpoint feasibility is entirely carried by $\varphi(t)$. Derivatives are
\begin{align*}
\dot{\phi}_n(t) = \tfrac{n\pi}{T}\cos\!\left(\tfrac{n\pi t}{T}\right), \quad
\ddot{\phi}_n(t) = -\left(\tfrac{n\pi}{T}\right)^2\sin\!\left(\tfrac{n\pi t}{T}\right).
\end{align*}

\subsubsection{Chebyshev basis}
\label{sec:chebyshev}
For polynomial representations, time is mapped to $x(t)=2t/T-1\in[-1,1]$, and the basis is given by Chebyshev polynomials of the first kind
\begin{equation}
    \begin{gathered}
        \phi_n(t) =  \mathcal{T}_n(x(t)), \\ 
         \mathcal T_n(x)=\cos(n\arccos x), \quad n=0,\dots,N .
    \end{gathered}    
\end{equation}
They are orthogonal on $[-1,1]$ with respect to the weight $(1-x^2)^{-\frac{1}{2}}$ and are evaluated via the three-term recurrence. First and second time derivatives are obtained by propagating $\frac{d\mathcal T_n}{dx}$ and $\frac{d^2 \mathcal T_n}{dx^2}$ and applying the chain rule with $\frac{dx}{dt}=\frac{2}{T}$.

\subsubsection{Quadratic regularization}
For all bases, a block-diagonal quadratic form $\mathbf Q=\mathbf I_{M \times M}\otimes\mathrm{diag}(w_0,\ldots,w_N)$ is constructed in coefficient space, with $w_n\propto n^2$, penalizing high-frequency components. Since $\bm\xi(t)$, $\dot{\bm\xi}(t)$, and $\ddot{\bm\xi}(t)$ depend linearly on $\bm\psi$ through \eqref{eq:traj}–\eqref{eq:traj_acc}, smoothness costs, dynamic constraints, and equality projections remain linear or quadratic in $\bm\psi$, preserving the sparse Gauss–Newton structure of \facto in the chosen function space.

\subsection{Single arm: cup carrying}
Since the cup-carrying task requires the cup axis aligned with the world's vertical, the task requirements in \eqref{eq:task_require} become
\begin{equation} \label{eq:cup_carry_task}
    \bm{\chi} = [\bm x_{\min}\quad \bm x_{\max}] = 
    \begin{bmatrix}
        -\pi & \pi \\
        \frac{\pi}{2} -\epsilon_\tau & \frac{\pi}{2} +\epsilon_\tau \\
        -\pi & \pi \\
        -\infty & \infty \\
        -\infty & \infty \\
        -\infty & \infty
    \end{bmatrix}. 
\end{equation}

Then we just need to encode it in the $\mathfrak{se}(3)$ and extract the $K'$ task constraints with significant task error.

\subsection{Dual arms: closed-chain collaboration on a rigid object.}

\begin{figure}[!htp]
    \centering
    \includegraphics[width=0.85\linewidth]{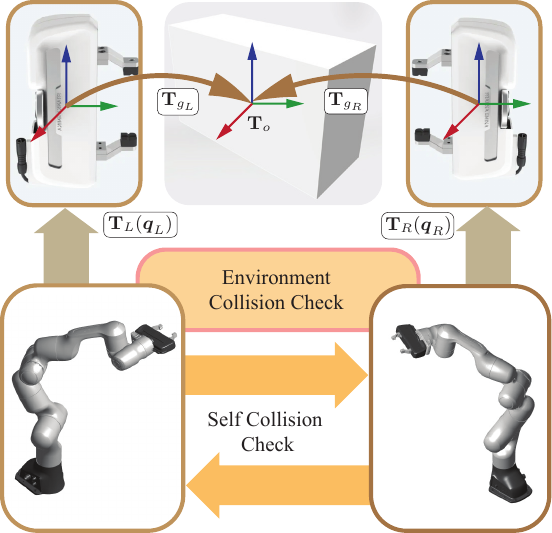}
    \caption{Dual-arm motion planning pipeline. Robot configurations $q_L$ and $q_R$ are mapped to end-effector poses $T_L(q_L)$ and $T_R(q_R)$. Candidate grasps, $T_{g_L}$ and $T_{g_R}$, are validated using self- and environment-collision checks.}
    \label{fig:dual-arm}
\end{figure}

We consider a dual-arm system, shown in \figref{fig:dual-arm}, with joint states $\bm q_L(t)\!\in\!\mathbb R^{M_L}$ and $\bm q_R(t)\!\in\!\mathbb R^{M_R}$, stacked as
\begin{equation}
    \bm q(t) = [\bm q_L^\top(t),\, \bm q_R^\top(t)]^\top \in \mathbb R^{M}, 
    \quad 
    M = M_L+M_R .
\end{equation}
Let the end–effector forward kinematics be $\mathbf T_L(\bm q_L),\mathbf T_R(\bm q_R)\in\mathrm{SE}(3)$.
Under rigid grasping, the two grippers must maintain a fixed \emph{relative transform}. Rather than explicitly estimating an object pose, we enforce the closed chain by constraining the \emph{relative motion} on $\mathrm{SE}(3)$.

\subsubsection{Relative transform on $\mathrm{SE}(3)$}

Define the instantaneous relative pose
\begin{equation}\label{eq:TRL_def}
    \mathbf T_{RL}(t) \triangleq \mathbf T_L^{-1}(\bm q_L(t))\,\mathbf T_R(\bm q_R(t)) \in \mathrm{SE}(3),
\end{equation}
and its reference value at the initial configuration $\bm q_0$,
\begin{equation}\label{eq:TRL_ref}
    \mathbf T_{RL}^0 \triangleq \mathbf T_L^{-1}(\bm q_{L,0})\,\mathbf T_R(\bm q_{R,0}) .
\end{equation}
Rigid closed-chain collaboration corresponds to $\mathbf T_{RL}(t)=\mathbf T_{RL}^0$ for all $t$, expressed in Lie algebra via logarithm map
\begin{equation}\label{eq:se3_err}
    \bm \xi(t) \triangleq \Log\!\left(\mathbf T_{RL}^{0^{-1}}\mathbf T_{RL}(t)\right) \in \mathfrak{se}(3),
    \qquad \bm \xi(t)=\bm 0 .
\end{equation}

\subsubsection{Implementation-aligned split: distance and relative posture}

In our implementation, we separate the constraint into (i) a \emph{relative distance} term that preserves the magnitude of the translational part, and (ii) an optional \emph{relative posture} term that additionally regulates the relative rotation and the displacement expressed in a shared frame.
Let $\mathbf d(t)$ denote the translation component of $\mathbf T_{RL}(t)$, i.e., $\mathbf T_{RL}(t) = (\mathbf R_{RL}(t),\,\mathbf d(t))$, and let $d_0=\|\mathbf d_0\|$ with $\mathbf d_0$ from $\mathbf T_{RL}^0$. The primary closed-chain condition is enforced as
\begin{equation}\label{eq:dist_err}
    e_{\text{pos}}(t) \;=\; \|\mathbf d(t)\| - \|\mathbf d_0\| \;=\; 0,
\end{equation}
which corresponds to preserving the inter–gripper distance.

When the relative posture constraint is enabled, we additionally regulate the rotational component 
$\mathbf R_{RL}(t)$ and the displacement expressed in a common reference frame. 
We construct a \emph{center} orientation $\mathbf R_c(t)\in\mathrm{SO}(3)$ from the two end–effector frames using the group geometric mean
\begin{equation}\label{eq:center_rot}
    \mathbf R_c(t) \triangleq \mathbf R_L(t)\, \Big(\mathbf R_L^\top(t)\mathbf R_R(t)\Big)^{1/2},
\end{equation}
and define the posture error as
\begin{equation} \label{eq:posture_err}
    \begin{aligned}
        \bm e_{\text{post}}(t)
        & \;=\;
        w_p\Big\|
            \mathbf R_c^\top(t)\mathbf d(t) - \mathbf R_c^{0^\top}\mathbf d_0
        \Big\|^2 \\
        & \;+\;
        w_R\big\|\Log\!\big((\mathbf R_{RL}^{0})^{-1} \mathbf R_{RL}(t)\big)\big\|^2,
    \end{aligned}
\end{equation}
where the second term is the $\mathfrak{so}(3)$ logarithm (equivalently the relative rotation angle), and $R_c^0$ is computed at $\bm q_0$. 

In this way, the collaboration task constraint becomes $\bm h_k = [\bm e_\text{pos}^\top(t_k), \bm e_\text{post}^\top(t_k)]^\top$ and it can be resolved by \facto (\algref{alg:facto}) based on the constraint linearization \eqref{eq:constr_stack}.

\section{Numerical Evaluation}
\label{sec:experiments}

In this section, the performance of \facto is first evaluated against six representative motion planners for single-arm manipulation across 800 unconstrained tasks provided by the motion bench maker (MBM \cite{chamzas_motionbenchmaker_2022}): CHOMP~\cite{zucker_chomp_2013}, TrajOpt~\cite{schulman_trust_2015}, GPMP2~\cite{mukadam_continuous-time_2018}, RRT-Connect (RRT-C \cite{kuffner_rrt-connect_2000}), RRT* \cite{karaman_sampling-based_2011}, and PRM~\cite{kavraki_analysis_1998}. For constrained manipulation, the framework is benchmarked against CHOMP and Atlas-RRT-Connect (Atlas-RRT-C \cite{jaillet_path_2013}) across 200 tasks. We further validate the approach's scalability using a dual-arm Franka robot across 40 unconstrained and 40 constrained scenarios. These benchmarks specifically focus on cluttered environments to assess continuous-time safety and dynamic smoothness. 
Finally, physical experiments demonstrate the practical utility of \facto for single-arm tasks and dual-arm coordination.

\subsection{Metrics}
\label{subsec:metrics}
We report the following metrics with mean and maximum values over $N=100$ trials per scene:
\begin{itemize}
\item \textbf{Success rate} (\%): fraction of collision-free trials\footnote{In our feasibility checks, we don't consider the collision detection between the fingers and the environment, because at the initial and goal configurations of some tasks provided by MBM, the fingers are attached to the object they tend to grasp. }.  
\item \textbf{Computation time} (s): wall-clock from problem instantiation to feasible trajectory, including the scaling of execution time $T$ (\secref{sec:time-scaling-sampled}). 
\item \textbf{Trajectory roughness}: roughness measured via integrated squared acceleration with a normalized $T = 1.0$:  
\begin{align*}
    \emph{roughness}  
    &= \frac{1}{K-1} \sum_{k=1}^{K-1}  \frac{\|\bm\theta_{k-1} - 2 \bm\theta_{k} + \bm\theta_{k+1}\|}{(t_k - t_{k-1})^2} \\
    &= (K-1) \sum_{k=1}^{K-1} \|\bm\theta_{k-1} - 2 \bm\theta_{k} + \bm\theta_{k+1}\|,
\end{align*}
where $\bm\theta_k = \bm\Psi(t_k) \bm\psi$ and $t_k = {k \cdot T}/{(K-1)}$. 
\end{itemize}

\subsection{Planner Configuration and Evaluation Protocol}

All planners were executed on a MacBook Pro equipped with a 2.3 GHz Intel Core i9 processor (single-threaded) and a 16 GB 2667 MHz DDR4 RAM. 
Considering fairness in comparisons, we use \textsc{Pinocchio} \cite{carpentier_pinocchio_2019} for kinematic calculations across all algorithms we want to benchmark. 
For the optimization-based algorithms, we use the same sphere-tree model and safety margin for collision check in a precomputed signed distance field \cite{osher_signed_2003}. Specifically, we reproduce the \textsc{MoveIt}-CHOMP  \cite{noauthor_moveit_nodate}, the \textsc{Tesseract}-TrajOpt \cite{noauthor_trajopt_2026} and GPMP2 using the \textsc{Pinocchio} \cite{carpentier_pinocchio_2019} and \textsc{OSQP} \cite{stellato_osqp_2020} libraries. 
For the sampling-based algorithms, we replace the validation module of \textsc{OMPL} \cite{sucan_open_2012} with \textsc{Pinocchio-FCL} for the mesh-pair collision query. 
In addition, IMACS \cite{kingston_exploring_2019}, using a projection-based constrained space with a constraint Jacobian, is applied to the constrained sampling. The planning time for all planners is capped at 10\,s, and memory usage is limited to 4\,GB.

\subsection{Parameter setting}

\tabref{tab:facto_params} lists the parameters used for \facto in all benchmark scenarios. The same parameter set is used across trials within each scenario. In addition, we test the \facto with 3 types of basis functions (sine, cosine, and Chebyshev) to demonstrate how our novel trajectory optimization framework, which searches for feasible coefficient solutions in a truncated sub-function space, performs across different scenarios.  These 3 types of \facto algorithms (\facto-Cheby, \facto-Sin, \facto-Cos) share the same parameter setting. 

\begin{table}[htp]
\centering
\caption{Parameter settings of \facto for benchmark scenarios.}
\begin{tabular}{lcccc}
\toprule
\textbf{Scenario} & \textbf{$\epsilon$} & \textbf{$N$} & \textbf{$\varrho$} & \textbf{$(\beta_1,\, \beta_2)$} \\
\midrule
bookshelf\_small     & 0.065 & 6 & 5e-2 & (0.250, 0.125) \\
bookshelf\_tall      & 0.065 & 6 & 5e-2 & (0.250, 0.125) \\
bookshelf\_thin      & 0.065 & 6 & 5e-2 & (0.250, 0.125) \\
box                  & 0.075 & 6 & 5e-2 & (0.250, 0.125) \\
kitchen              & 0.075 & 7 & 5e-2 & (0.250, 0.125) \\
cage                 & 0.035 & 8 & 5e-3 & (0.500, 0.250) \\
table\_under\_pick   & 0.065 & 6 & 5e-3 & (0.250, 0.125) \\
table\_pick          & 0.065 & 6 & 5e-2 & (0.250, 0.125) \\
\midrule
kitchen\_constr      & 0.075 & 7 & 5e-2 & (0.250, 0.125) \\
table\_constr        & 0.065 & 7 & 5e-3 & (0.250, 0.125) \\
\midrule
table\_dual\_arm          & 0.075 & 6 & 1e-2 & (0.250, 0.125) \\
table\_dual\_arm\_constr  & 0.075 & 7 & 1e-2 & (0.500, 0.250) \\
\bottomrule
\end{tabular}
\label{tab:facto_params}
\end{table}

The obstacle clearance parameter $\epsilon$ specifies the minimum safety margin enforced during optimization. For most scenes, a nominal value of $\epsilon = 0.065$--$0.075$ provided stable convergence. In the \emph{cage} environment, the margin was reduced to $\epsilon = 0.035$ to accommodate narrow passages.
The number of Fourier bases $N$ determines the trajectory representation capacity. A default value of $N=6$ was sufficient for open and moderately cluttered scenes. More constrained environments required higher expressiveness, with $N=7$ for \emph{kitchen} and $N=8$ for \emph{cage}.
The smoothness weight $\varrho$ penalizes high-frequency components in the trajectory coefficients. A value of $\varrho = 5\times10^{-2}$ was used in most scenarios, while a smaller weight $\varrho = 5\times10^{-3}$ was applied in confined settings to allow finer local adjustments.
The EMA weights $(\beta_1,\beta_2)$ smooth out the uneven updates to the obstacle function. These are fixed to $(0.25,0.125)$ for all scenes except \emph{cage}, where larger values $(0.50,0.25)$ improve feasibility under tight constraints.

The number of discrete waypoints for CHOMP and TrajOpt is fixed as 40 to be consistent with the maximum number of collision-check nodes for \facto. 
As for GPMP2, the number of support waypoints matches the number of \facto basis functions, and the number of up-sampled intermediate waypoints is fixed at 5. 
Moreover, their safety margin $\epsilon$ for collision-check and the smoothness weight $\varrho$ are consistent with \facto's. 
We also finely tune some key parameters, such as the \texttt{learning\_rate} of CHOMP and GPMP2, and the \texttt{init\_trust\_region\_size} of TrajOpt, to ensure good performance for the algorithms we have reproduced. 
All other parameters are set to their default values. 

For the sampling-based motion planners (RRT-C, RRT*, PRM, and IMACS-RRT-C), we use their default parameters, set the maximum computation-time budget as 10.0 s, and opt out of any path simplification method. Considering the validity of the start and goal waypoints, we also delete the meshes of the left and right fingers in the bookshelf and cage scenarios.


\begin{table*}[!htp]
\centering
\caption{Single-arm benchmark results of the unconstrained tasks in \emph{bookshelf} scenarios.}
\label{tab:bookshelf}
\begin{tabular}{llccccccccc}
\toprule
Scene & Metric & \facto-Cheby & \facto-Sin & \facto-Cos & CHOMP & TrajOpt & GPMP2 & RRT-C & RRT* & PRM \\
\midrule
\multirow{5}{*}{\emph{bookshelf\_small}}
 & Success (\%) & 93 & 97 & 98 & 98 & 67 & 97 & 96 & 47 & 78 \\
 & Time (s) & 0.0376 & 0.2114 & 0.0387 & 0.1010 & 0.1297 & 0.0624 & 0.6753 & 1.5649 & 0.8481 \\
 & Time max (s) & 0.1819 & 0.5763 & 0.1235 & 0.2410 & 0.4280 & 0.1594 & 9.5129 & 9.9939 & 8.4673 \\
 & Roughness & 52.41 & 17.32 & 52.02 & 36.59 & 96.17 & 22.52 & 75.15 & 61.16 & 178.60 \\
 & Roughness max & 100.31 & 66.30 & 85.62 & 54.06 & 579.64 & 37.42 & 555.34 & 198.49 & 408.42 \\
\midrule
\multirow{5}{*}{\emph{bookshelf\_tall}}
 & Success (\%) & 98 & 98 & 98 & 98 & 84 & 97 & 93 & 28 & 83 \\
 & Time (s) & 0.0260 & 0.2272 & 0.0272 & 0.0886 & 0.1348 & 0.0631 & 0.4828 & 1.4760 & 1.5758 \\
 & Time max (s) & 0.0922 & 0.6732 & 0.0785 & 0.2347 & 0.3479 & 0.2592 & 8.0935 & 9.3306 & 9.2312 \\
 & Roughness & 55.10 & 16.33 & 40.97 & 37.47 & 61.78 & 22.52 & 79.65 & 56.34 & 116.58 \\
 & Roughness Max & 100.07 & 47.35 & 85.75 & 59.75 & 235.97 & 37.42 & 455.57 & 172.87 & 553.57 \\
\midrule
\multirow{5}{*}{\emph{bookshelf\_thin}}
 & Success (\%) & 99 & 100 & 100 & 100 & 63 & 94 & 97 & 25 & 61 \\
 & Time (s) & 0.0457 & 0.1965 & 0.0478 & 0.0926 & 0.1426 & 0.0666 & 0.8085 & 3.2522 & 2.2643 \\
 & Time max (s) & 0.1210 & 0.7410 & 0.1043 & 0.2170 & 0.4900 & 0.1422 & 7.6820 & 9.6943 & 10.3143 \\
 & Roughness & 49.97 & 15.80 & 39.11 & 34.67 & 65.64 & 21.93 & 92.78 & 65.60 & 116.88 \\
 & Roughness max & 85.73 & 48.14 & 76.45 & 50.60 & 297.89 & 37.95 & 532.78 & 169.41 & 536.14 \\
\bottomrule
\end{tabular}
\end{table*}

\begin{table*}[!htp]
\centering
\caption{Single-arm benchmark results of the unconstrained tasks in \emph{box} and \emph{cage} scenarios.}
\label{tab:box_cage}
\begin{tabular}{llccccccccc}
\toprule
Scene & Metric & \facto-Cheby & \facto-Sin & \facto-Cos & CHOMP & TrajOpt & GPMP2 & RRT-C & RRT* & PRM \\
\midrule
\multirow{5}{*}{\emph{box}}
 & Success (\%) & 90 & 92 & 98 & 98 & 22 & 93 & 99 & 4 & 85 \\
 & Time (s) & 0.0838 & 0.1430 & 0.0546 & 0.0961 & 0.2428 & 0.0590 & 0.2927 & 4.7499 & 1.9173 \\
 & Time max (s) & 0.3000 & 0.4993 & 0.2863 & 0.2390 & 0.4431 & 0.3306 & 4.4057 & 9.4519 & 9.5398 \\
 & Roughness & 40.91 & 18.43 & 53.14 & 33.94 & 225.98 & 22.48 & 120.63 & 68.01 & 178.07 \\
 & Roughness max & 78.90 & 92.20 & 103.93 & 46.99 & 472.49 & 40.59 & 466.65 & 139.62 & 443.24 \\
\midrule
\multirow{5}{*}{\emph{cage}}
 & Success (\%) & 95 & 93 & 95 & 19 & 0 & 82 & 92 & 0 & 17 \\
 & Time (s) & 0.0794 & 0.3058 & 0.1345 & 0.1545 & -- & 0.2133 & 3.7793 & -- & 4.5376 \\
 & Time max (s) & 0.0544 & 0.7128 & 1.3377 & 0.2347 & -- & 1.5455 & 10.0057 & -- & 9.9040 \\
 & Roughness & 76.92 & 26.10 & 91.06 & 47.85 & -- & 48.22 & 386.26 & -- & 139.45 \\
 & Roughness max & 155.36 & 61.98 & 154.50 & 55.34 & -- & 107.44 & 1322.45 & -- & 435.91 \\
\bottomrule
\end{tabular}
\end{table*}

\begin{table*}[!htp]
\centering
\caption{Single-arm benchmark results of the unconstrained tasks in \emph{kitchen} and \emph{table\_pick} scenarios.}
\label{tab:kitchen_table}
\begin{tabular}{llccccccccc}
\toprule
Scene & Metric & \facto-Cheby & \facto-Sin & \facto-Cos & CHOMP & TrajOpt & GPMP2 & RRT-C & RRT* & PRM \\
\midrule
\multirow{5}{*}{kitchen}
 & Success (\%) & 87 & 90 & 94 & 85 & 43 & 74 & 99 & 55 & 78 \\
 & Time (s) & 0.1044 & 0.1710 & 0.0841 & 0.1592 & 0.2247 & 0.1126 & 0.5383 & 1.6377 & 1.2332 \\
 & Time max (s) & 0.3939 & 0.5831 & 0.9290 & 0.4375 & 0.5261 & 0.1998 & 6.0240 & 8.7961 & 8.2058 \\
 & Roughness & 81.45 & 41.94 & 79.26 & 53.69 & 299.30 & 33.01 & 189.48 & 149.74 & 274.39 \\
 & Roughness max & 133.54 & 106.45 & 154.38 & 115.46 & 212.52 & 46.36 & 824.69 & 506.71 & 630.49 \\
\midrule
\multirow{5}{*}{\emph{table\_pick}}
 & Success (\%) & 91 & 98 & 99 & 87 & 56 & 97 & 98 & 33 & 85 \\
 & Time (s) & 0.0633 & 0.3375 & 0.0421 & 0.1310 & 0.1517 & 0.0264 & 0.2306 & 2.9695 & 1.3845 \\
 & Time max (s) & 0.2597 & 0.6141 & 0.1033 & 0.3211 & 0.4177 & 0.2437 & 4.4521 & 9.8163 & 7.4833 \\
 & Roughness & 43.48 & 16.33 & 55.64 & 37.56 & 93.60 & 22.42 & 65.06 & 46.74 & 148.34 \\
 & Roughness max & 84.93 & 49.68 & 106.45 & 53.73 & 285.39 & 32.81 & 345.16 & 131.10 & 360.64 \\
\midrule
\multirow{5}{*}{\emph{table\_under\_pick}}
 & Success (\%) & 69 & 87 & 98 & 36 & 56 & 83 & 98 & 16 & 84 \\
 & Time (s) & 0.0611 & 0.2244 & 0.0595 & 0.1588 & 0.2081 & 0.0906 & 0.2377 & 2.2044 & 1.8546 \\
 & Time max (s) & 0.2627 & 0.5974 & 0.2065 & 0.2366 & 0.4361 & 0.1840 & 5.6697 & 9.8101 & 10.0026 \\
 & Roughness & 210.35 & 40.15 & 85.40 & 45.78 & 340.49 & 34.31 & 144.66 & 138.89 & 134.11 \\
 & Roughness max & 380.34 & 115.31 & 137.13 & 63.45 & 771.20 & 53.78 & 377.17 & 298.94 & 553.52 \\
\bottomrule
\end{tabular}
\end{table*}

\begin{figure*}[!htp]
  \centering
  \hfill
  \begin{subfigure}[t]{1\linewidth}
      \includegraphics[width=0.16\linewidth]{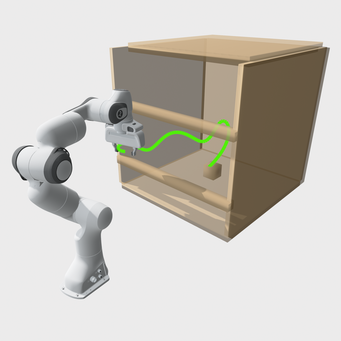}
      \includegraphics[width=0.16\linewidth]{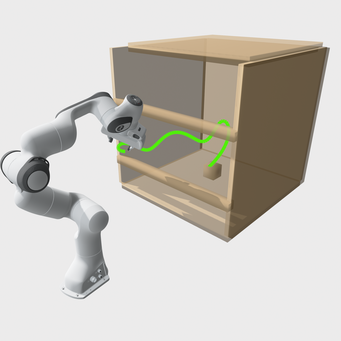}
      \includegraphics[width=0.16\linewidth]{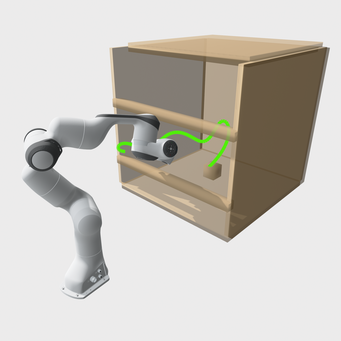}
      \includegraphics[width=0.16\linewidth]{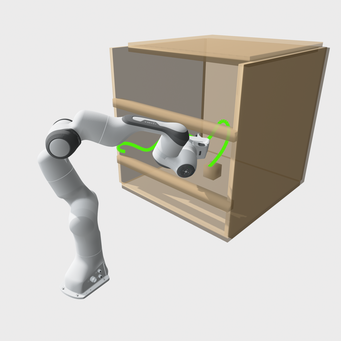}
      \includegraphics[width=0.16\linewidth]{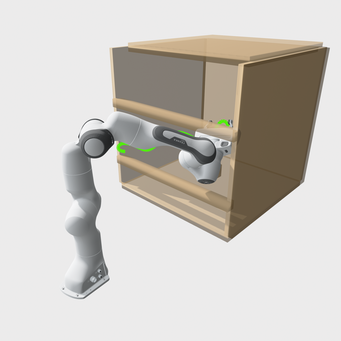}
      \includegraphics[width=0.16\linewidth]{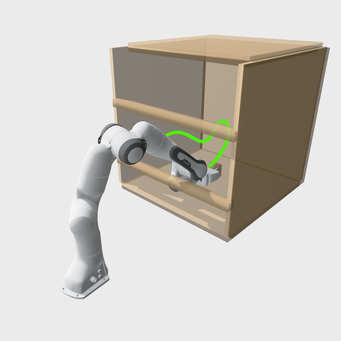}
      \caption{\emph{cage} scenario (task-unconstrained): the end-effector is free of task constraints. The robot plans a collision-free reaching motion inside the box-like cage; representative snapshots along the trajectory are shown (green).}
  \end{subfigure}
  \hfill
  \caption{Result of cage scenario, single arm for unconstrained tasks}
  \label{fig:result_box_cage}
\end{figure*}
\begin{figure*}[!htp]
  \centering
  \hfill
  \begin{subfigure}[t]{1\linewidth}
    \centering
    \includegraphics[width=0.16\linewidth]{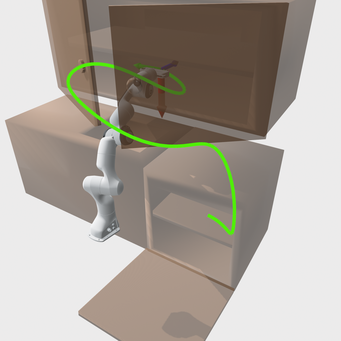}
    \includegraphics[width=0.16\linewidth]{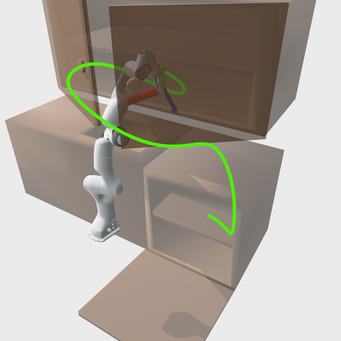}
    \includegraphics[width=0.16\linewidth]{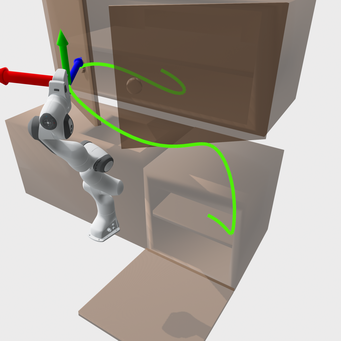}
    \includegraphics[width=0.16\linewidth]{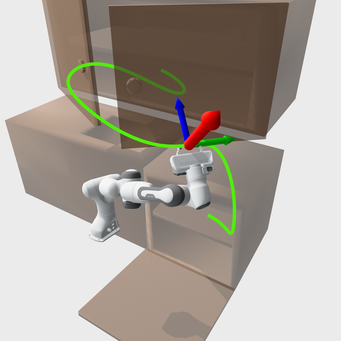}
    \includegraphics[width=0.16\linewidth]{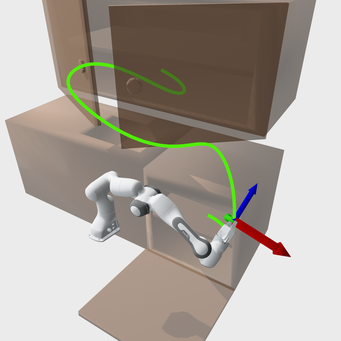}
    \includegraphics[width=0.16\linewidth]{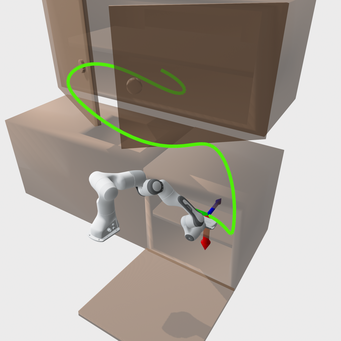}
    \caption{\emph{kitchen} scenario (unconstrained): the end-effector is free of task constraints. The robot plans a collision-free reaching motion in a cluttered kitchen setting; representative snapshots along the trajectory are shown (green).}
  \end{subfigure}
  \hfill
  \\[2pt]
  \hfill
  \begin{subfigure}[t]{1\linewidth}
    \centering
    \includegraphics[width=0.16\linewidth]{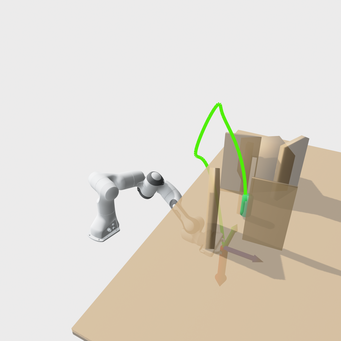}
    \includegraphics[width=0.16\linewidth]{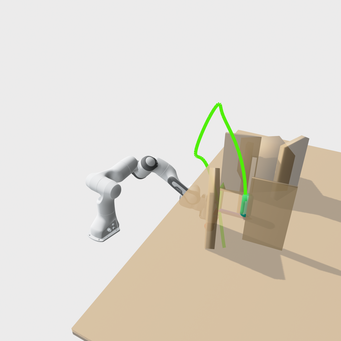}
    \includegraphics[width=0.16\linewidth]{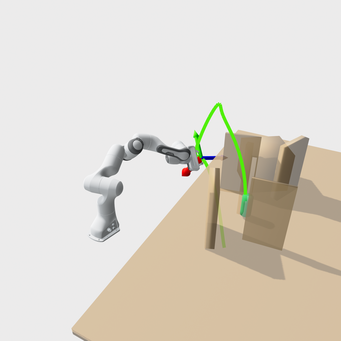}
    \includegraphics[width=0.16\linewidth]{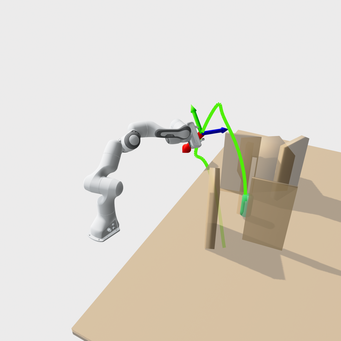}
    \includegraphics[width=0.16\linewidth]{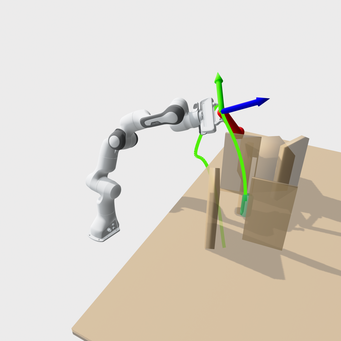}
    \includegraphics[width=0.16\linewidth]{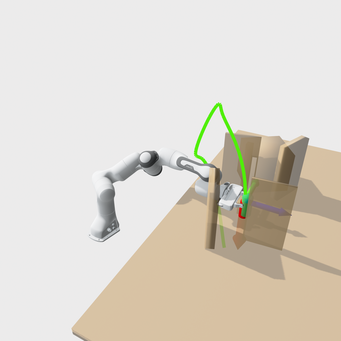}
    \caption{\emph{table\_under\_pick} scenario (unconstrained): the end-effector is free of task constraints. The robot plans a collision-free reaching motion under the table and around obstacles; representative snapshots along the trajectory are shown (green).}
  \end{subfigure}
  \hfill
  \caption{Single-arm result of the unconstrained tasks in \emph{kitchen} and \emph{table\_under\_pick} scenarios}
  \label{fig:result_kitchen_table}
\end{figure*}
%

\subsection{Benchmark for Single-arm Tasks}

\subsubsection{Result analysis for the unconstrained cases}
\label{sec:result_single_arm_unconstr}

\tabref{tab:bookshelf}, \tabref{tab:box_cage}, and \tabref{tab:kitchen_table} summarize single-arm \emph{unconstrained} planning in cluttered scenes. The main difficulty remains narrow passages (\figref{fig:result_box_cage}) and steep collision gradients (\figref{fig:result_kitchen_table}). These effects show up in feasibility and roughness.

\paragraph{Comparison across algorithms}

For planning reliability within a limited time budget, \facto-Cos achieves the highest success rate of 94\%-100\%, which is even higher than RRT-C's 92\%-99\%. For computational efficiency, the average running time of optimization-based planners (0.03\,s-0.3\,s) is much faster than that of sampling-based planners (0.2\,s-4.7\,s).  In terms of trajectory smoothness, the average roughness of optimization-based planners is lower than that of sampling-based planners. 

More specifically, \facto-Sin achieves the lowest roughness of 15-42 due to its straight-line initial guess, but it takes the highest computation cost in the \facto family. Though the success rate of CHOMP and GPMP2 is lower than \facto-Sin, they achieve sub-second smoothness with higher efficiency. That is because they both directly solve the objective, penalize constraint violations, and use line search rather than the constrained SQP by OSQP. This is also evident in the efficiency comparison between CHOMP/GPMP2 and TrajOpt-OSQP. GPMP2 is more reliable than CHOMP/TrajOpt because it parameterizes the trajectory space with a GP, rather than directly employing discrete waypoints as CHOMP/TrajOpt does. Moreover, RRT-C achieves the highest efficiency and success rate among sampling-based planners by bidirectional search without rewiring. PRM gains sub-second performance, and RRT* achieves the lowest roughness.

\paragraph{Comparison across scenarios}

The planning difficulty increases as the number of obstacles surrounding the feasible solutions increases, the solutions approach the joint limits, and the solutions move farther from the initial guess. Therefore, we sort the difficulty of the 8 scenarios as \emph{cage} > \emph{table\_under\_pick} = \emph{kitchen} > \emph{box} > \emph{table\_pick} > \emph{bookshelf}, which is consistent with the success rate. 

In the 3 \emph{bookshelf} scenes, most optimization-based planners, except for TrajOpt, succeed in almost 100\% of planning tasks, and the computation time of \facto-Cheby/Cos, and GPMP2 achieves $10^{-2}$\,s, showing their high reliability and efficiency over the sampling-based planners. However, CHOMP drops to 19\% and 36\% in \emph{cage} and \emph{table\_under\_pick}, and GPMP2 drops to 74 \% in \emph{kitchen}, but \facto-Cos still keeps high (94\%-98\%) in these scenes. For the sampling-based planner, RRT-C remains reliable and highly efficient, with only a slight performance decline in \emph{cage}, where RRT* and PRM even drop to 0\% and 17\%.

\paragraph{Effect of function-space basis within \facto}

In the \facto family, \facto-Cheby/Cos achieves the highest efficiency, meanwhile \facto-Sin has the lowest roughness. Though \facto-Cheby performs more efficiently in \emph{bookshelf} and \emph{cage} scenes, \facto-Cos is still the most reliable across all scenes, even better than RRT-C.

\begin{table*}[!htp]
\centering
\caption{Single-arm benchmark results for task-constrained scenarios.}
\label{tab:task_constrained_single_arm}
\begin{tabular}{llccccc}
\toprule
Scene & Metric & FACTO-Cheby & FACTO-Sin & FACTO-Cos & CHOMP & IMACS-RRT-C \\
\midrule
\multirow{5}{*}{\emph{kitchen\_constr}}
 & Success (\%) & 79 & 88 & 90 & 10 & 86 \\
 & Time (s) & 0.1800 & 0.5602 & 0.2077 & 1.5915 & 4.0364 \\
 & Time max (s) & 0.8345 & 1.6304 & 1.0268 & 1.8074 & 8.7817 \\
 & Roughness & 125.12 & 93.97 & 141.20 & 335.49 & 2117.17 \\
 & Roughness max & 482.55 & 266.14 & 424.33 & 1852.22 & 7570.24 \\
\midrule
\multirow{5}{*}{\emph{table\_under\_pick\_constr}}
 & Success (\%) & 77 & 88 & 96 & 9 & 72 \\
 & Time (s) & 0.1140 & 0.6585 & 0.1768 & 1.4922 & 4.1853 \\
 & Time max (s) & 0.5816 & 1.2603 & 0.9838 & 1.8341 & 10.0562 \\
 & Roughness & 169.83 & 79.69 & 177.56 & 347.06 & 2297.13 \\
 & Roughness max & 400.36 & 385.99 & 472.06 & 2043.64 & 5229.90 \\
\bottomrule
\end{tabular}
\end{table*}
\begin{figure*}[!htp]
  \centering
  \hfill
  \begin{subfigure}[t]{1\linewidth}
    \centering
    \includegraphics[width=0.16\linewidth]{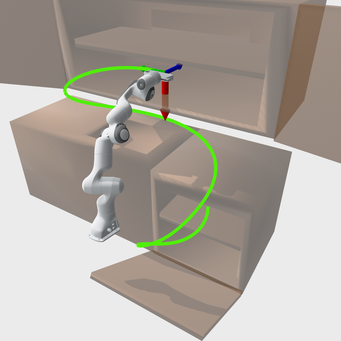}
    \includegraphics[width=0.16\linewidth]{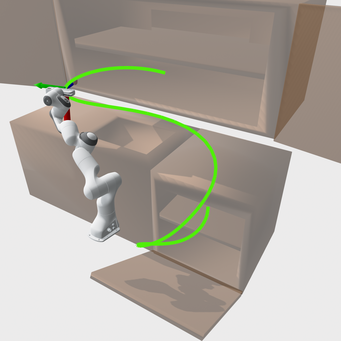}
    \includegraphics[width=0.16\linewidth]{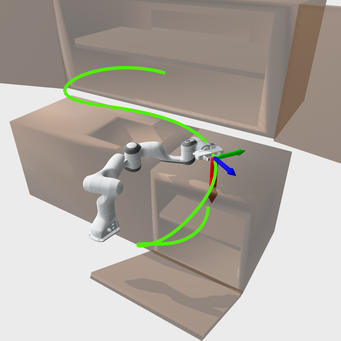}
    \includegraphics[width=0.16\linewidth]{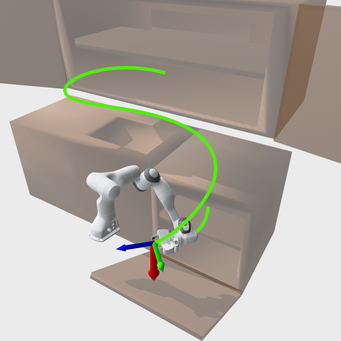}
    \includegraphics[width=0.16\linewidth]{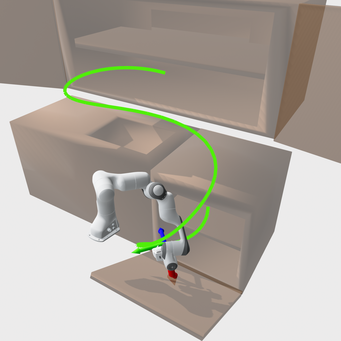}
    \includegraphics[width=0.16\linewidth]{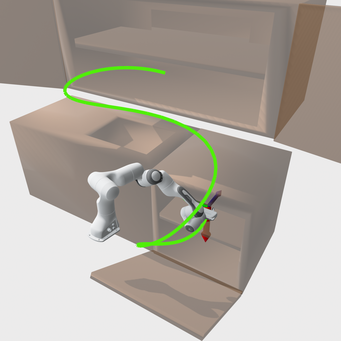}
    \caption{\emph{kitchen} scenario (constrained): the end-effector $x$-axis (red arrow) is constrained to align with the world $z$-axis. The robot plans a collision-free reaching motion while maintaining the orientation constraint; representative snapshots along the trajectory are shown (green). }
    \end{subfigure}
  \hfill
  \\[2pt]
  \hfill
  \begin{subfigure}[t]{1\linewidth}
    \centering
    \includegraphics[width=0.16\linewidth]{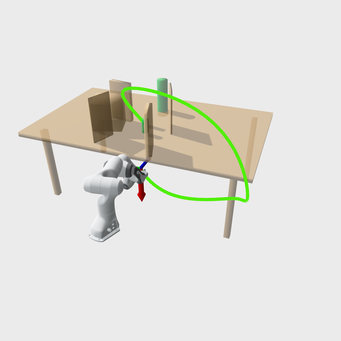}
    \includegraphics[width=0.16\linewidth]{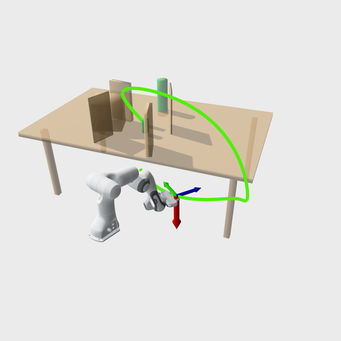}
    \includegraphics[width=0.16\linewidth]{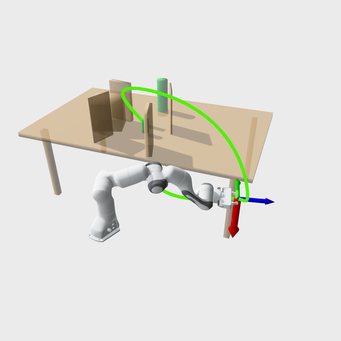}
    \includegraphics[width=0.16\linewidth]{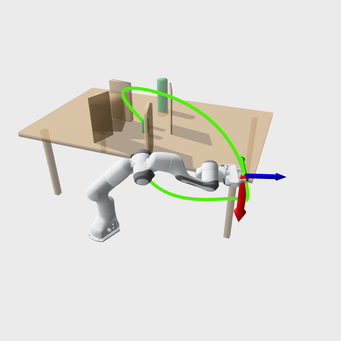}
    \includegraphics[width=0.16\linewidth]{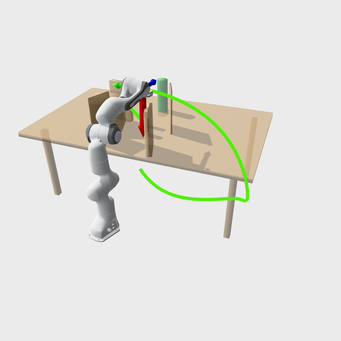}
    \includegraphics[width=0.16\linewidth]{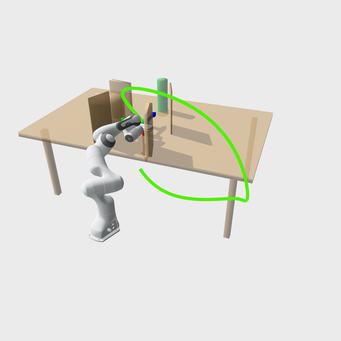}
    \caption{\emph{table\_under\_pick} scenario (constrained): the end-effector $x$-axis (red arrow) is constrained to align with the world $z$-axis. The robot plans a collision-free reaching motion around the table and obstacles while maintaining the orientation constraint; representative snapshots along the trajectory are shown (green).}
  \end{subfigure}
  \hfill
  \caption{Single-arm results of the constrained tasks in \emph{kitchen} and \emph{table} scenarios}
  \label{fig:result_kitchen_table_constr}
\end{figure*}

\subsubsection{Result analysis for the constrained cases}\label{sec:constrained}

\tabref{tab:task_constrained_single_arm} reports two task-constrained single-arm scenes, shown in \figref{fig:result_kitchen_table_constr}. The constraint collapses the feasible set to a thin manifold. Small collision perturbations matter more. Curvature also grows quickly after projection or recovery steps.

\paragraph{Comparison across algorithms}

For planning reliability, \facto-Cos achieves the highest success rate (90\% \& 96\%), \facto-Cheby offers the best computational efficiency (0.18\,s \& 0.11\,s on average), and \facto-Sin yields the smoothest results (94 \& 80 on average) in terms of roughness. Though IMACS-RRT-C has the third success rate (86\% \& 72 \%), its computational cost is around $\times 20$ higher than \facto-Cos/Cheby since it uses the tangent bundle for almost every sample in the IMACS framework. CHOMP performs worst in both reliability and efficiency.

\paragraph{Constrained v.s. unconstrained tasks}

The success rate of all planners declines except for \facto-Sin in \emph{table\_under\_pick} from the unconstrained task (\figref{fig:result_kitchen_table}) to the constrained task (\figref{fig:result_kitchen_table_constr}) due to the restricted feasible space. Meanwhile, computational costs increase because the task constraint requires more numerical projections, thereby introducing greater search fluctuations for both convergence and sampling processes.

\paragraph{Basis effect within \facto}

Though \facto-Sin has the lowest roughness and \facto-Cheby has the smallest computation cost, \facto-Cos still has the best balance among reliability, efficiency, and smoothness.

\subsection{Dual-Arm Motion Planning}

All planners were evaluated within a unified dual-arm planning framework. The experiments use two 7-DoF \textsc{Franka Panda} manipulators operating in a shared workspace. 
Both arms were modeled in a Pinocchio kinematic tree, with collision checking handled by FCL. 

\begin{table*}[!htp]
\centering
\caption{Benchmark results for unconstrained dual-arm coordination scenario.}
\label{tab:dualarm_coord_unconstr}
\begin{tabular}{llccccccccc}
\toprule
Scene & Metric & FACTO-Cheby & FACTO-Sin & FACTO-Cos & CHOMP & TrajOpt & GPMP2 & RRT-C & RRT* & PRM \\
\midrule
\multirow{5}{*}{\emph{table\_under\_pick}}
 & Success (\%) & 70 & 90 & 92.5 & 27.5 & 27.5 & 47.5 & 90 & 12.5 & 12.5 \\
 & Time (s) & 0.2056 & 0.4776 & 0.2421 & 0.4379 & 0.4767 & 0.3116 & 2.0272 & 7.0996 & 5.3386 \\
 & Time max (s) & 0.3828 & 1.0061 & 0.8534 & 0.5005 & 0.5297 & 0.8220 & 8.3906 & 8.1623 & 10.0060 \\
 & Roughness & 134.59 & 80.76 & 250.56 & 88.85 & 88.85 & 60.00 & 512.03 & 337.10 & 498.88 \\
 & Roughness max & 331.68 & 176.61 & 466.41 & 108.28 & 108.28 & 88.56 & 1013.21 & 628.82 & 847.33 \\
\bottomrule
\end{tabular}
\end{table*}
\begin{table*}[!htp]
\centering
\caption{Benchmark results for constrained dual-arm coordination scenario.}
\label{tab:dualarm_coord_constr}
\begin{tabular}{llccccc}
\toprule
Scene & Metric & FACTO-Cheby & FACTO-Sin & FACTO-Cos & CHOMP & IMACS-RRT-C \\
\midrule
\multirow{5}{*}{\emph{table\_under\_pick\_constr}}
 & Success (\%) & 62.5 & 77.5 & 87.5 & 7.5 & 27.5 \\
 & Time (s) & 0.2453 & 1.5373 & 0.3703 & 3.5326 & 4.2324 \\
 & Time max (s) & 0.3948 & 2.6837 & 0.6560 & 4.2109 & 8.6093 \\
 & Roughness & 197.39 & 129.74 & 322.18 & 95.55 & 1122.33 \\
 & Roughness max & 647.02 & 270.11 & 488.10 & 140.23 & 2674.44 \\
\bottomrule
\end{tabular}
\end{table*}
\begin{figure*}[!htp]
  \centering
  \hfill
  \begin{subfigure}[t]{1\linewidth}
      \includegraphics[width=0.16\linewidth]{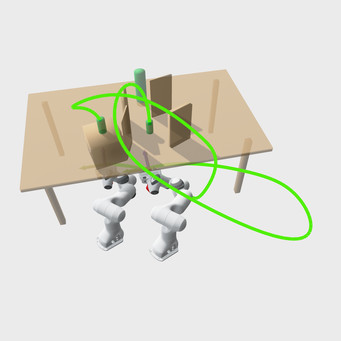}
      \includegraphics[width=0.16\linewidth]{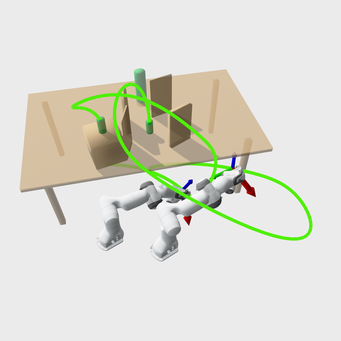}
      \includegraphics[width=0.16\linewidth]{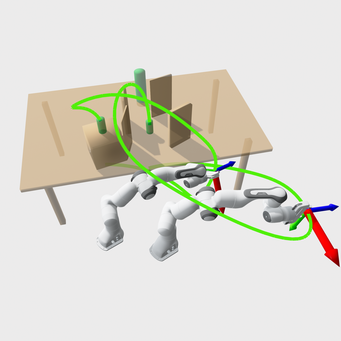}
      \includegraphics[width=0.16\linewidth]{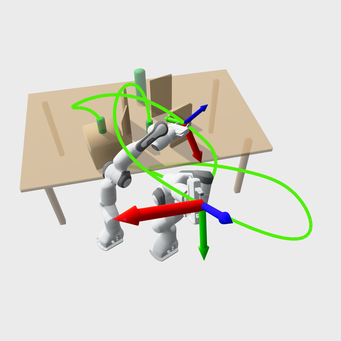}
      \includegraphics[width=0.16\linewidth]{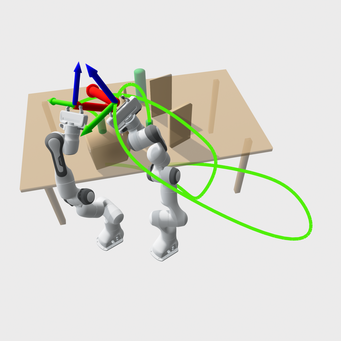}
      \includegraphics[width=0.16\linewidth]{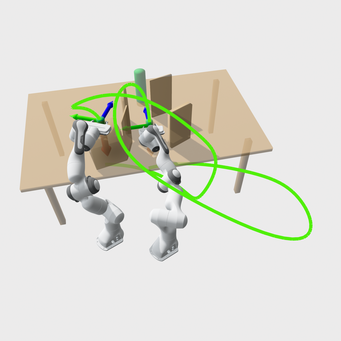}
      \caption{\emph{table\_under\_pick} scenario (dual-arm, unconstrained): the end-effectors are free of task constraints. The two arms plan a coordinated collision-free reaching motion around the table and obstacles; representative snapshots along the trajectory are shown (green). }
      \label{fig:result_table_under_pick_dual_arm}
  \end{subfigure}
  \hfill
  \\[2pt]
  \hfill
  \begin{subfigure}[t]{1\linewidth}
      \includegraphics[width=0.16\linewidth]{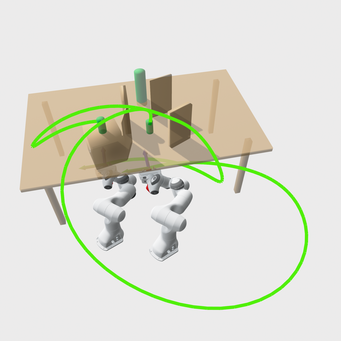}
      \includegraphics[width=0.16\linewidth]{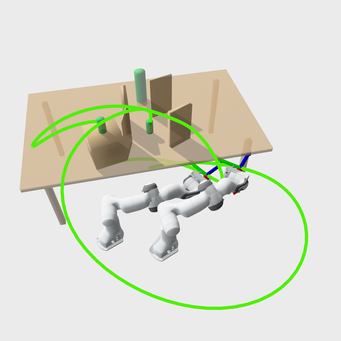}
      \includegraphics[width=0.16\linewidth]{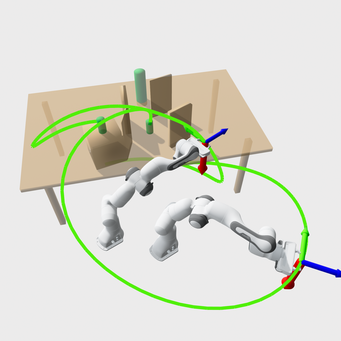}
      \includegraphics[width=0.16\linewidth]{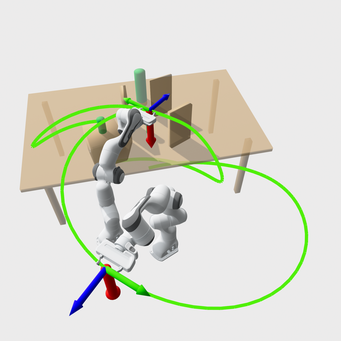}
      \includegraphics[width=0.16\linewidth]{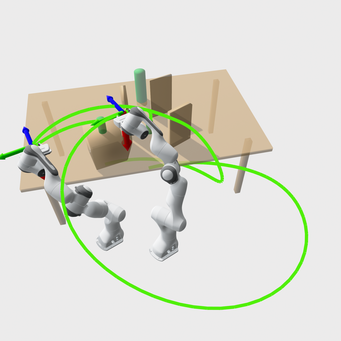}
      \includegraphics[width=0.16\linewidth]{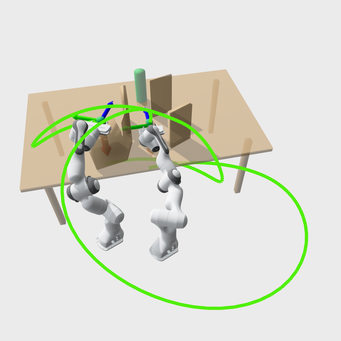}
      \caption{\emph{table\_under\_pick} scenario (dual-arm, constrained): the end-effectors $x$-axes (red arrows) are constrained to align with the world $z$-axis. The two arms plan a coordinated collision-free reaching motion while maintaining the orientation constraints; representative snapshots along the trajectory are shown (green).}
      \label{fig:result_table_under_pick_constr_dual_arm}
  \end{subfigure}
  \hfill
  \caption{Results of dual-arm coordination in a table scenario}
\end{figure*}

\subsubsection{Setup}

The benchmark includes 40 trials per scenario, covering an unconstrained reach task (\figref{fig:result_table_under_pick_dual_arm}) and a task-constrained variant (\figref{fig:result_table_under_pick_constr_dual_arm}). All planners used identical environment meshes and start/goal configurations. Compared to single-arm planning, the dual-arm setting introduces stronger coupling and narrower feasible regions due to inter-arm interaction. The \facto parameter settings follow \tabref{tab:facto_params}.

\subsubsection{Result analysis}

Tables \ref{tab:dualarm_coord_unconstr}\,\&\,\ref{tab:dualarm_coord_constr} evaluate dual-arm coordination in a \textsc{table\_under\_pick} scene, without and with a task constraint. In the unconstrained case, collision complexity and inter-arm coupling dominate. The constrained case adds a manifold requirement. That further reduces the feasible set and makes optimization steps more aggressive. Roughness grows quickly when feasibility is recovered by projection.

\paragraph{Comparison across algorithms}

In the unconstrained case, \facto-Cos achieves the highest success rate of 92.5\%, followed by RRT-C and \facto-Sin (90\%). However, RRT-C takes $\times 5 \sim \times 10$ computation cost compared to \facto-Sin/Cos. GPMP2 has the third-highest success rate of 47.5\%, with a low computational cost (0.3\,s on average). Its slightly higher average computation time is due to up-sampling. CHOMP and TrajOpt perform worse because they use discrete waypoints for trajectory representation, which leads to between-waypoint safety issues. RRT* and PRM perform worst within the 10\,s time budget because their rewiring and query/connection processes require excessive computation, particularly in the dual-arm case with 14 dimensions.  

In the constrained case, \facto-Cos has a significant lead over the others with a success rate of 87.5\%. Though \facto-Cheby's computational time (0.2\,s) and \facto-Sin's roughness (80) are less than \facto-Cos's (0.4\,s, 322), their success rate (77.5\%, 62.5\%) is much lower. IMACS-RRT-C's reliability follows the \facto family, but its success rate is still below 50\% with $\times 10$ computation cost and roughness. CHOMP performs too poorly to implement. 

\paragraph{Constrained v.s. unconstrained tasks}

As in the single-arm case, all planners' success rates of the dual-arm case decline as the feasible space shrinks. The \facto-family drops slightly from 70\%-92.5\% to 62.5\%-87.5\%, while CHOMP and IMACS-RRT-C drop significantly from 27.5\% \& 90\% to 7.5\% \& 27.5\%. The reason is also similar to the single-arm case: when the feasible workspace is highly constrained, CHOMP cannot handle between-waypoint safety, and IMAC-RRT-C cannot project samples onto the constraint manifold within a 10\,s time budget. Likewise, roughness and computational cost increase due to excessive manifold projection.

\section{Real-world Experiment Validation}
\label{sec:hardware_validation}

\subsection{System setup}
All physical experiments are conducted on FR3 manipulators using a real-time control stack for deterministic trajectory replay. The low-level controller runs on an ASUS NUC with Ubuntu~22.04 and a real-time (RT) kernel. The robots are connected via Ethernet using UDP, and all commands are issued through the Franka Control Interface (FCI) with libfranka~0.15.0. Collision behaviors and safety limits are kept fixed across the entire suite.


\begin{figure*}[!htp]
  \centering
  \hfill
  \begin{subfigure}[t]{1\linewidth}
      \centering
      \includegraphics[width=0.16\linewidth]{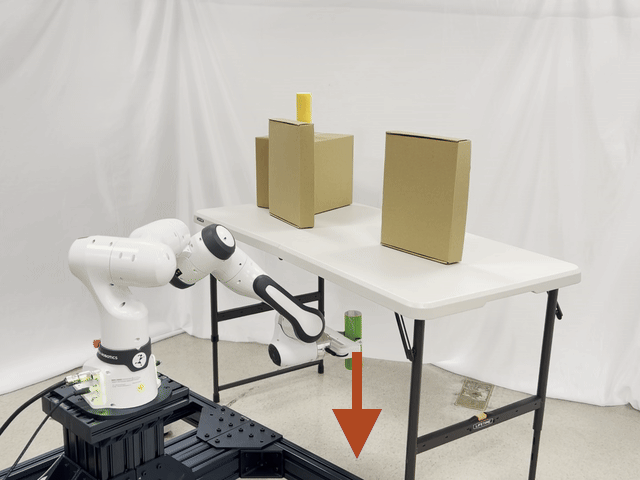}
      \includegraphics[width=0.16\linewidth]{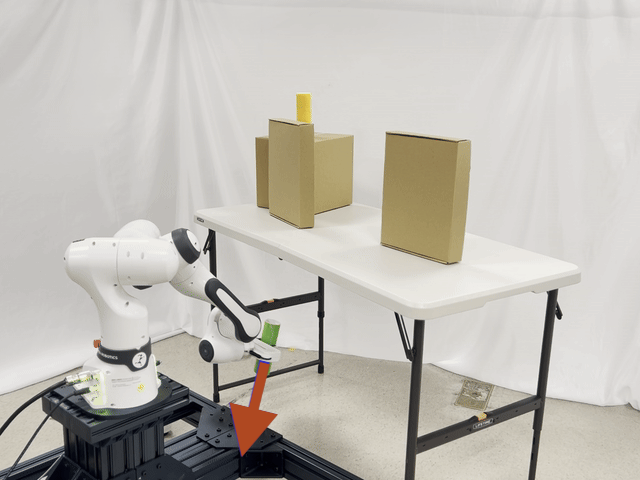}
      \includegraphics[width=0.16\linewidth]{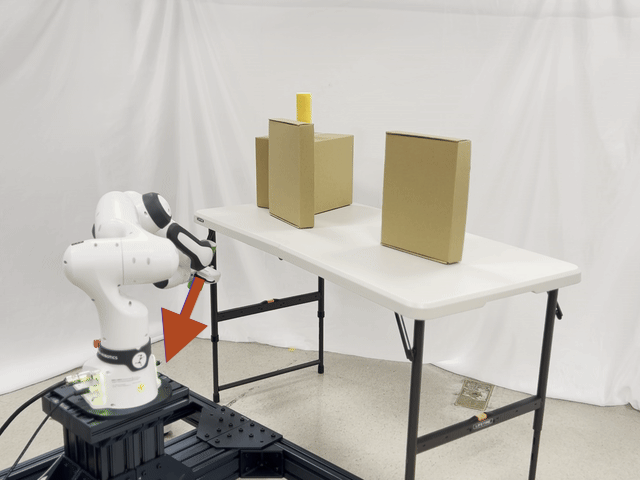}
      \includegraphics[width=0.16\linewidth]{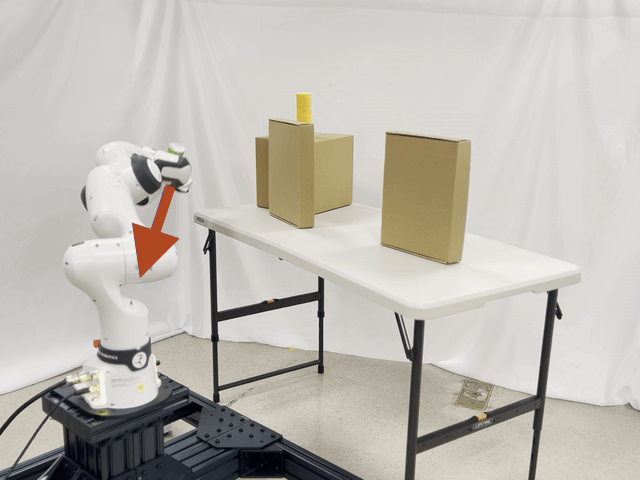}
      \includegraphics[width=0.16\linewidth]{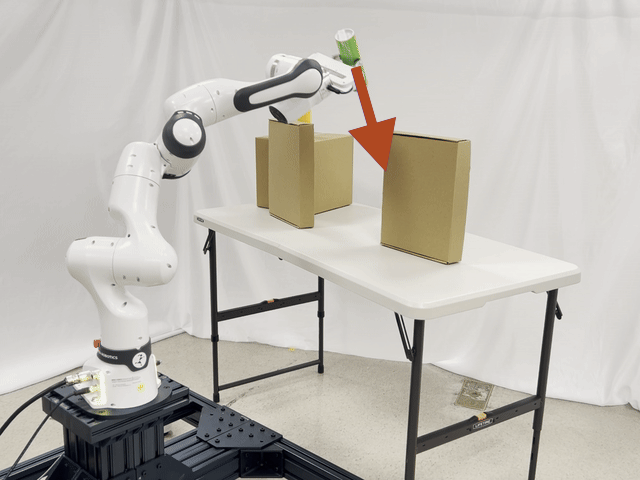}
      \includegraphics[width=0.16\linewidth]{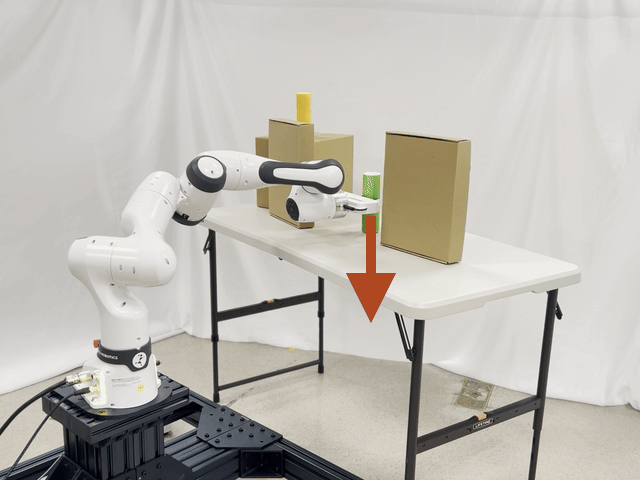}
      \caption{\emph{table\_under\_pick} scenario (hardware, unconstrained): the end-effector is free of task constraints. The FR3 executes a collision-free motion to retrieve a tube from under the table; representative snapshots along the trajectory are shown. }
      \label{fig:exp_table_under_pick}
  \end{subfigure}
  \\[2pt]
  \hfill
  \begin{subfigure}[t]{1\linewidth}
      \centering
      \includegraphics[width=0.16\linewidth]{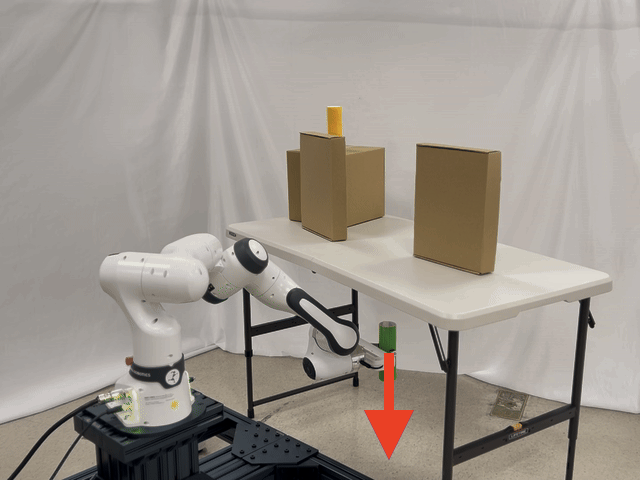}
      \includegraphics[width=0.16\linewidth]{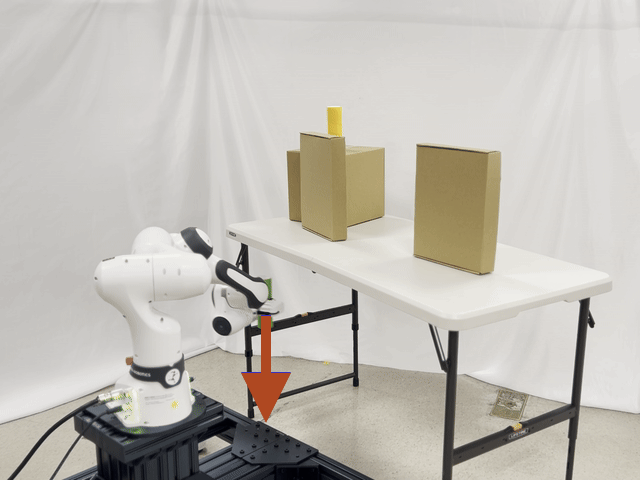}
      \includegraphics[width=0.16\linewidth]{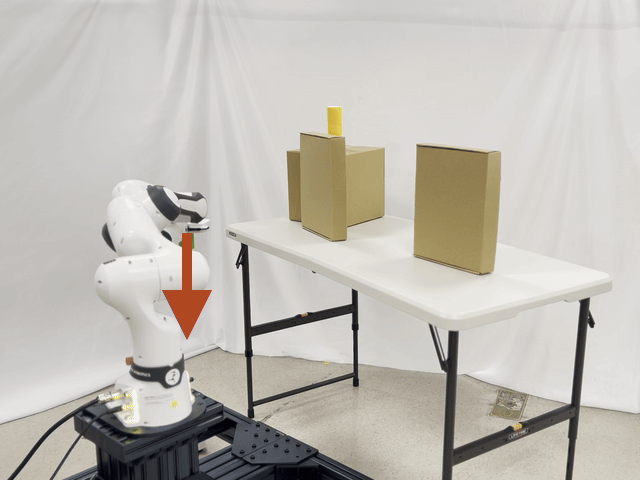}
      \includegraphics[width=0.16\linewidth]{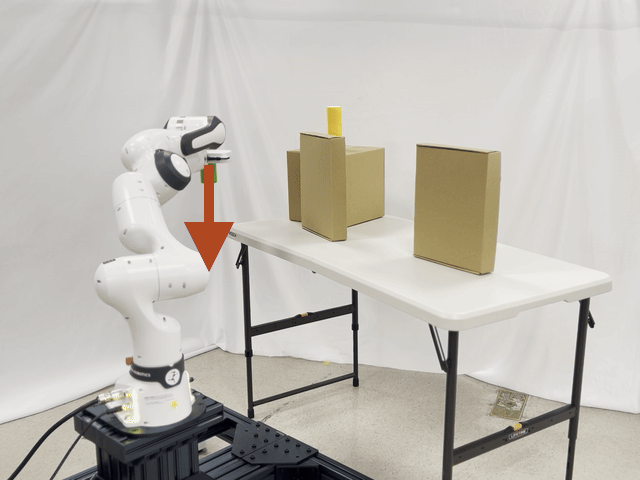}
      \includegraphics[width=0.16\linewidth]{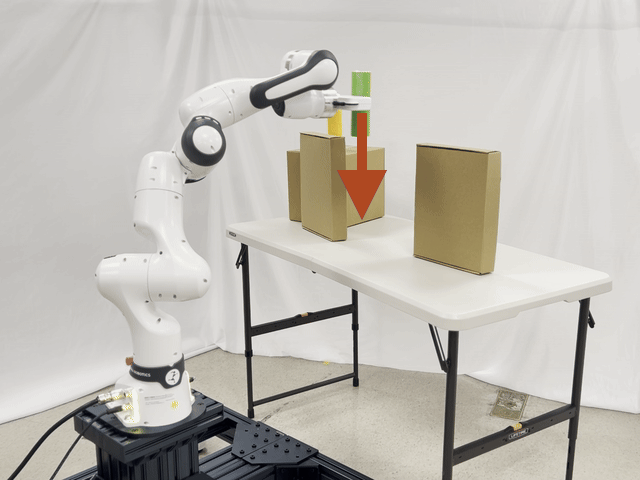}
      \includegraphics[width=0.16\linewidth]{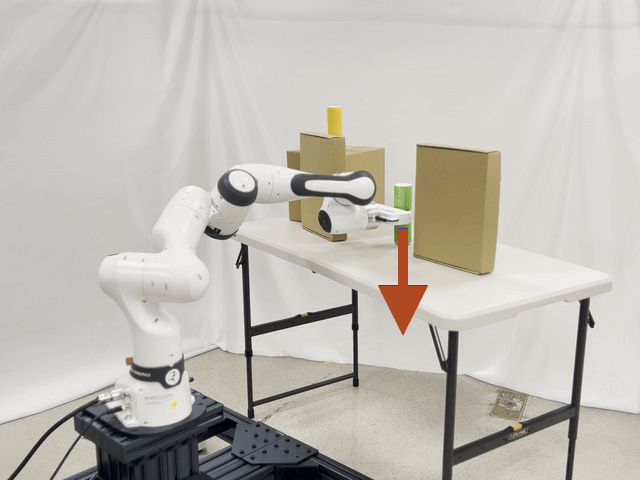}
      \caption{\emph{table\_under\_pick} scenario (hardware, constrained): the end-effector $x$-axis (red arrow) is constrained to align with the world $z$-axis. The FR3 retrieves a tube from under the table and places it on the tabletop while maintaining the pose constraint; representative snapshots along the execution are shown.}
      \label{fig:exp_table_under_pick_constr}
  \end{subfigure}
  \hfill
  \caption{Physical validation results of a single arm executing 2 tasks in the table scenario}
  \label{fig:exp_single_arm}
\end{figure*} 

\begin{figure*}[!htp]
  \centering
  \hfill
  \begin{subfigure}[t]{1\linewidth}
    \centering
    \includegraphics[width=0.16\linewidth]{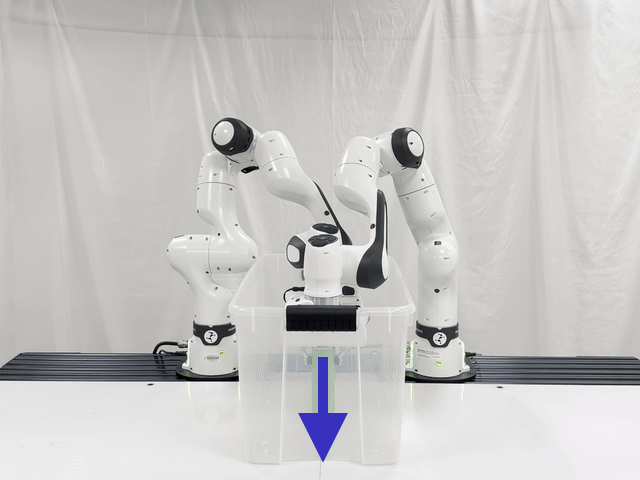}
    \includegraphics[width=0.16\linewidth]{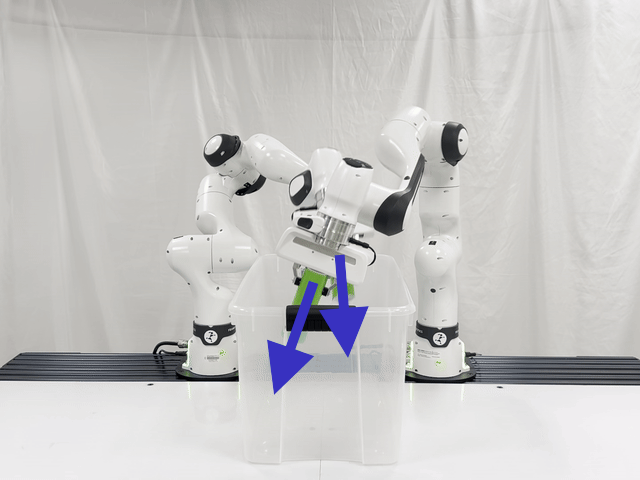}
    \includegraphics[width=0.16\linewidth]{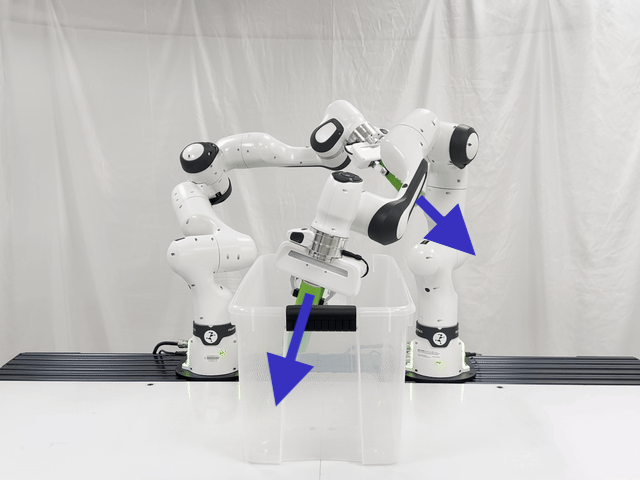}
    \includegraphics[width=0.16\linewidth]{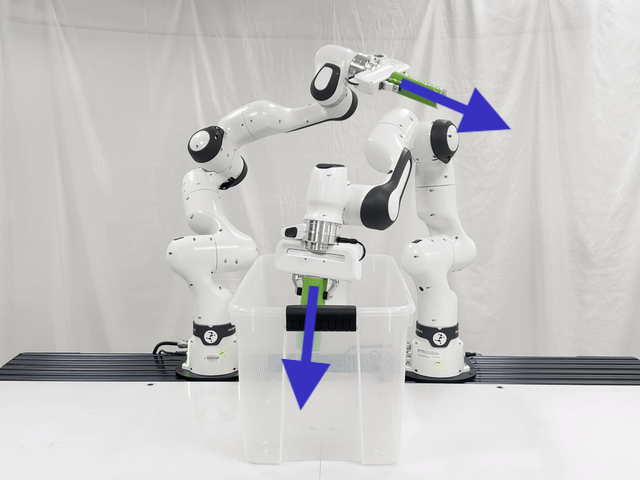}
    \includegraphics[width=0.16\linewidth]{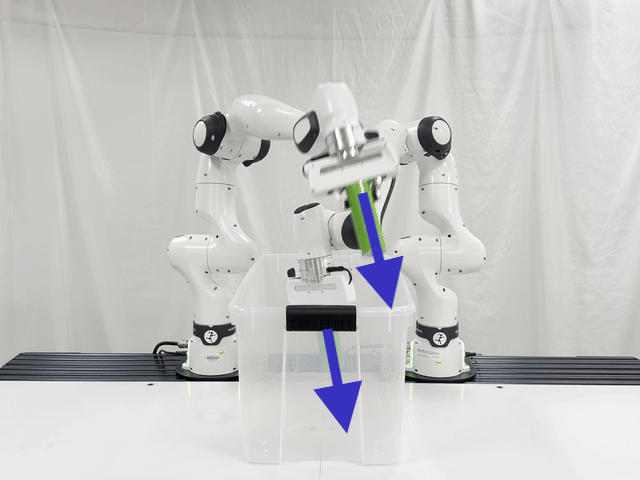}
    \includegraphics[width=0.16\linewidth]{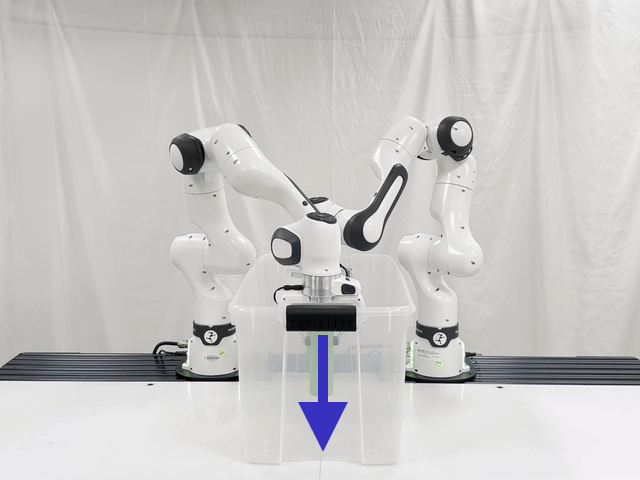}
    \caption{\emph{dual\_arm\_02} scenario (hardware, task-unconstrained): the end-effectors are free of task constraints. The two FR3 arms coordinate to swap the positions of two tubes; representative snapshots along the execution are shown.}
    \label{fig:exp_dual_arm_02}
  \end{subfigure}
  \hfill
  \\[2pt]
  \hfill
  \begin{subfigure}[t]{1\linewidth}
    \centering
    \includegraphics[width=0.16\linewidth]{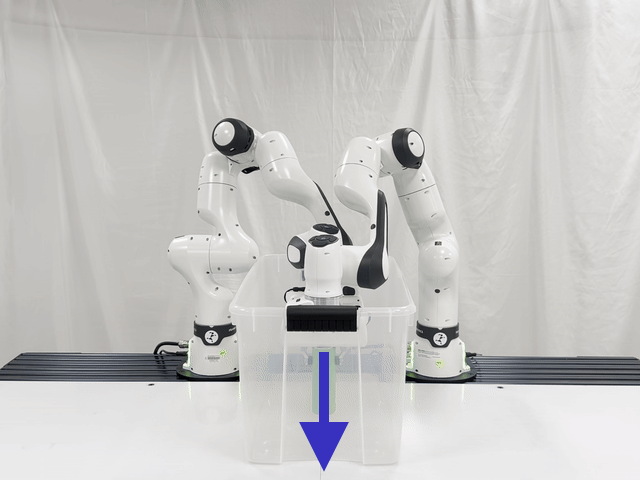}
    \includegraphics[width=0.16\linewidth]{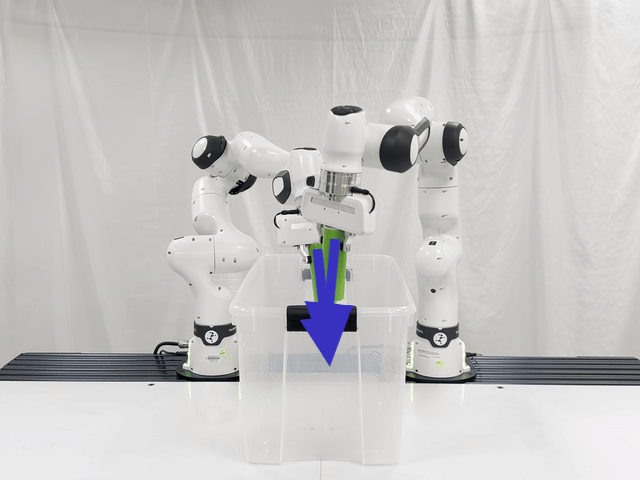}
    \includegraphics[width=0.16\linewidth]{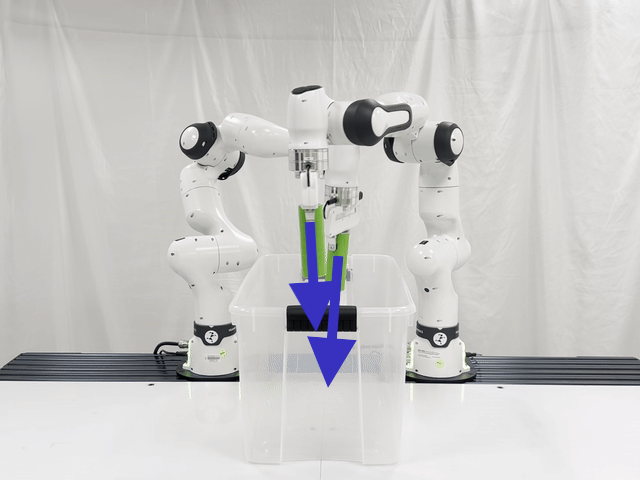}
    \includegraphics[width=0.16\linewidth]{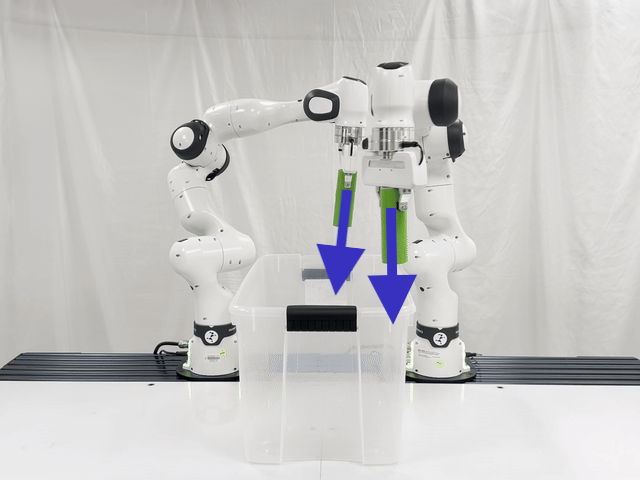}
    \includegraphics[width=0.16\linewidth]{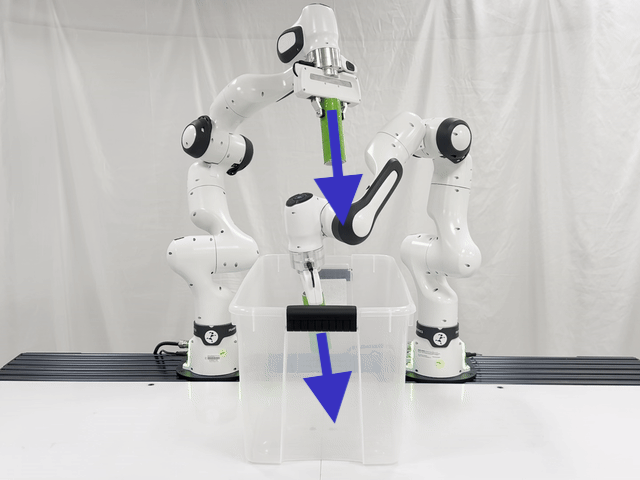}
    \includegraphics[width=0.16\linewidth]{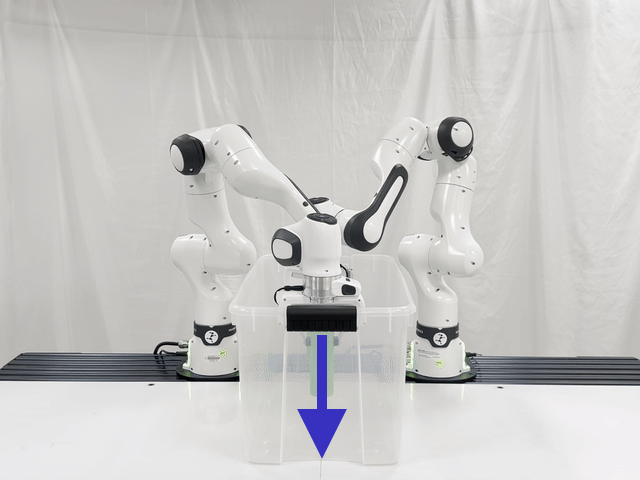}
    \caption{\emph{dual\_arm\_constr} scenario (hardware, task-constrained): the end-effector $z$-axes (blue arrows) are constrained to align with the world $z$-axis. The two FR3 arms swap the positions of two tubes while maintaining the orientation constraints; representative snapshots along the execution are shown.}
    \label{fig:exp_dual_arm_constr}
  \end{subfigure}
  \hfill
  \\[2pt]
  \hfill
  \begin{subfigure}[t]{1\linewidth}
    \centering
    \includegraphics[width=0.16\linewidth]{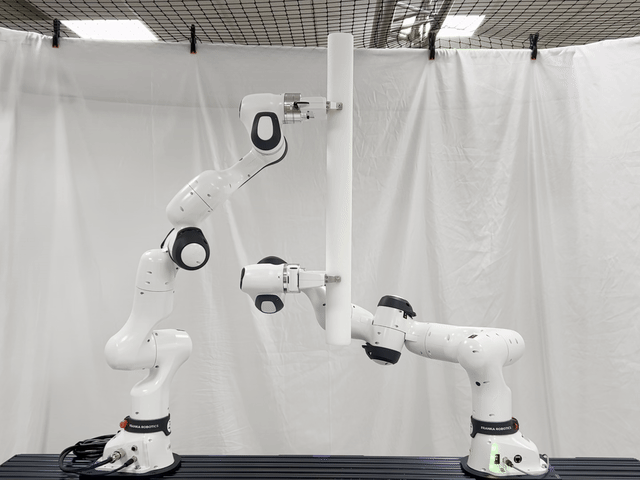}
    \includegraphics[width=0.16\linewidth]{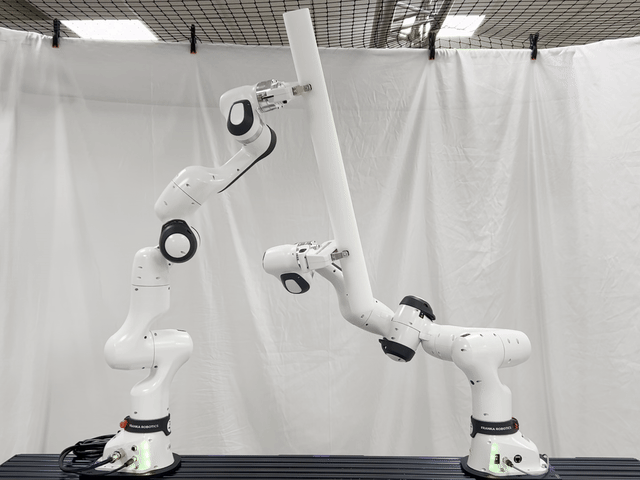}
    \includegraphics[width=0.16\linewidth]{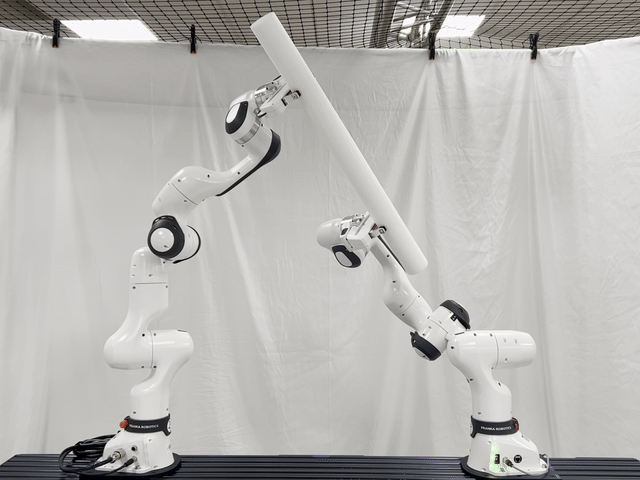}
    \includegraphics[width=0.16\linewidth]{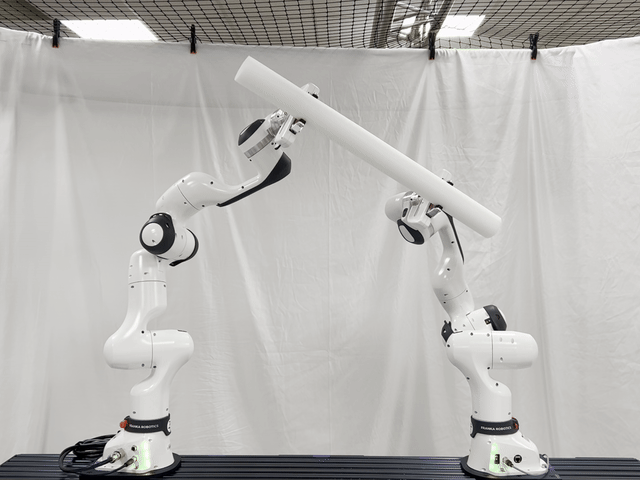}
    \includegraphics[width=0.16\linewidth]{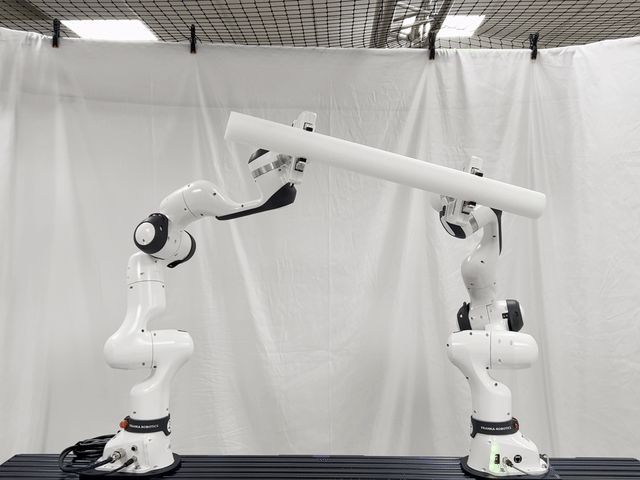}
    \includegraphics[width=0.16\linewidth]{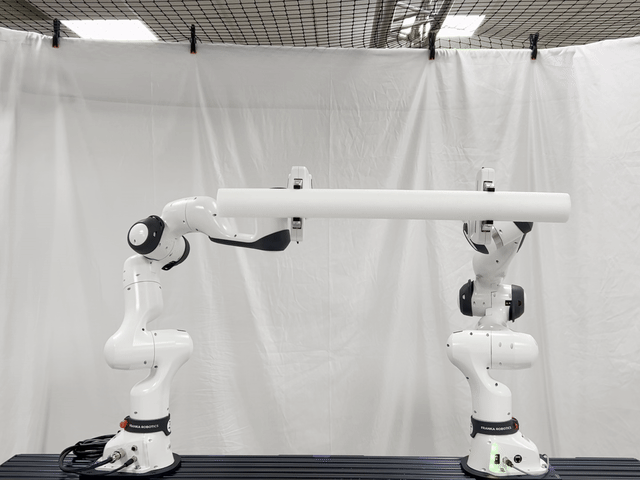}
    \caption{\emph{dual\_arm\_collab} scenario (hardware, task-unconstrained): the end-effectors are free of task constraints. The two FR3 arms collaboratively rotate a long roller by $90^\circ$; representative snapshots along the execution are shown.}
    \label{fig:exp_dual_arm_collab}
  \end{subfigure}
  \hfill
  \caption{Physical validation results of the dual arms executing 3 tasks in 2 scenarios}
  \label{fig:exp_dual_arm}
\end{figure*}

\subsection{Experiment setup}

To demonstrate the implementation potential of \facto, we validate 10 physical demonstrations, comprising 6 single-arm trials across 3 scenarios (bookshelf, box, and table) and 4 dual-arm trials across 2 scenarios (box and table). Specifically, Figures \ref{fig:exp_table_under_pick}--\ref{fig:exp_table_under_pick_constr} compare the validation results for the unconstrained and constrained single-arm trials using the same initial and goal waypoints in the same workspace. Figures \ref{fig:exp_dual_arm_02}--\ref{fig:exp_dual_arm_constr} do a similar comparison of the dual-arm coordination trails. Moreover, \figref{fig:exp_dual_arm_collab} validates the dual-arm collaboration. 

\subsection{Execution protocol}

At the beginning of trail testing, each robot is moved towards a pre-defined initial waypoint. Then each robot executes its own joint trajectory parameterized by the coefficients $\bm\psi$ over the scaled execution time period $T$, where the reference joint velocity is computed according to \eqref{eq:traj_vel} and joint acceleration is interpolated by libfranka. In the dual-arm trial, we execute each robot's trajectory in its own thread and subnet. 

\begin{table}[!htp]
\centering
\caption{Physical experiment suite and time-scaling settings.}
\label{tab:hardware_suite_times}
\begin{tabular}{llccc}
\toprule
{Robot} & {Scenario} & {$N$} & {Scaled time (s)} \\
\midrule
\multirow{6}{*}{{Single arm}} 
& \emph{bookshelf}                  & 6 & 2.754 \\
& \emph{bookshelf\_half}            & 6 & 3.517 \\
& \emph{box}                        & 6 & 4.626 \\
& \emph{table\_pick}                & 6 & 5.853 \\
& \emph{table\_under\_pick}         & 6 & 5.138 \\
& \emph{table\_under\_pick\_constr} & 7 & 5.291 \\
\midrule
\multirow{4}{*}{{Dual arms}} 
& \emph{dual\_arm\_01}              & 6 & 3.110 \\
& \emph{dual\_arm\_02}              & 6 & 3.285 \\
& \emph{dual\_arm\_constr}          & 7 & 6.479 \\
& \emph{dual\_arm\_collab}          & 8 & 4.084 \\
\bottomrule
\end{tabular}
\end{table}
%

\subsection{Efficiency and the effect of time-scaling}

\tabref{tab:hardware_suite_times} lists the number of basis functions $N$ and the scaled execution time of each robot trial, calculated by \facto-Cos. It shows that all the unconstrained trials use 6 basis functions for trajectory parameterization and their scaled time ranges from 2.7\,s to 5.8\,s. Meanwhile, the basis amount for all constrained trials exceeds 6 and, for the dual-arm collaboration, even 8. It is consistent with our earlier observation that more constraints lead to greater shrinkage of the feasible space, thereby requiring a larger $N$ to maintain feasibility. In corresponding, their scaled time also increases, ranging from 4.0\,s to 6.5\,s.


Overall, our physical experiment comprising 10 tasks across 6 scenarios for both single- and dual-arm systems demonstrates the high implementation potential of \facto. All trajectories, parameterized by the optimized coefficients, can be executed smoothly and efficiently in a physical workspace. It also validates the effectiveness of \facto's \texttt{TimeScalar} (cf. \algref{alg:time-scaling}).

\section{Conclusion and Discussion}
This paper presents \facto, a function-space adaptive constrained trajectory optimizer that parameterizes a function space of trajectories using finite coefficients as the coordinates of basis functions. It uses an EMA-smoothed, convexified objective to optimize the trajectory in the null space of the linearized constraint, with adaptive trust regions. The block structure is also proposed for the multi-robot system. 

The benchmark results on 800 single-arm unconstrained tasks validate \facto's high reliability relative to other optimization-based planners, such as CHOMP, TrajOpt, and GPMP2, and its high efficiency relative to sampling-based planners, such as RRT-Connect, RRT*, and PRM. These high performances are even more significant in 200 single-arm constrained tasks, showing \facto's handling of trajectory-wide constraints within a truncated coefficient space. The comparison between \facto and CHOMP/GPMP2 further validates the effectiveness of the EMA-based GN method for fast, stable convergence and demonstrates efficiency gains from variable reduction achieved by coefficient-space truncation. These well performances of \facto in 40 unconstrained and 40 constrained dual-arm tasks further certify \facto's solvability of the multi-robot problem. 

The physical experiment demonstrates \facto's implementation potential for single-arm manipulation and dual-arm coordination/collaboration tasks. It also validates the executability and smoothness of the well-parameterized trajectory under execution-time scaling, accounting for the velocity and torque limits of the FR3 arm.

\section{Limitation and Future Work}

Although the benchmark demonstrates the high reliability and efficiency of \facto relative to traditional optimization- and sampling-based motion planners, its per-step computational cost remains higher than that of CHOMP and GPMP2, but much lower than that of TrajOpt.  As we have analyzed the single-arm unconstrained benchmark results, it mainly comes from the QP solving procedure: CHOMP and GPMP2 directly penalize the avoidance constraints in their objectives and solve the QP problem via a single line-search step; however, \facto and TrajOpt directly solves the constrained QP via OSQP, a operater splitting method alternating between linear system solve and constraint projection frequently in one QP step. Therefore, our future work will focus on balancing these two methods, leveraging the high efficiency of the penalty method and the high feasibility of the operator-splitting method. 

Moreover, although \facto has the highest efficiency and reliability across 14 scenarios (12 single-arm and 2 dual-arm), its performance still declines as the number of arms, task difficulty, or the tightness of task constraints increases. Our future work will also integrate additional deep learning or reinforcement learning techniques to further improve the algorithmic robustness and adaptability to variant scenarios.

\bibliographystyle{IEEEtran}
\bibliography{reference}

\end{document}